\documentclass{article}

\usepackage[preprint]{neurips_2025}

\usepackage{enumitem}
\usepackage[hidelinks]{hyperref}
\usepackage{url}
\usepackage{booktabs}
\usepackage{amsfonts}
\usepackage{graphicx}
\usepackage{fontawesome}
\usepackage{sidecap}
\usepackage[small]{caption}
\usepackage{subcaption}
\usepackage{amsmath}
\allowdisplaybreaks
\usepackage{amsthm}

\usepackage{wrapfig}
\usepackage{booktabs}
\usepackage{multicol}

\usepackage{float}
\usepackage{xcolor}
\usepackage{colortbl}
\usepackage{amssymb}
\usepackage{adjustbox}
\usepackage{pifont}
\usepackage{enumitem}
\usepackage[noabbrev,capitalize]{cleveref}

\usepackage[compact]{titlesec}
\titlespacing{\section}{0pt}{1ex}{0.5ex}
\titlespacing{\subsection}{0pt}{0.5ex}{0ex}
\titlespacing{\subsubsection}{0pt}{0.5ex}{0ex} 
\usepackage{setspace}
\AtBeginDocument{%
  \addtolength\abovedisplayskip{-0.25\baselineskip}%
  \addtolength\belowdisplayskip{-0.25\baselineskip}%
  \addtolength\abovedisplayshortskip{-0.25\baselineskip}%
  \addtolength\belowdisplayshortskip{-0.25\baselineskip}%
}

\usepackage{pifont}
\usepackage{xspace}
\newcommand{\circone}{\ding{172}\xspace}
\newcommand{\circtwo}{\ding{173}\xspace}

\usepackage{tabularx, booktabs}
\newcolumntype{C}{>{\centering\arraybackslash}X}
\newcolumntype{R}{>{\raggedleft\arraybackslash}X}
\newcolumntype{S}{>{\raggedleft\arraybackslash\hsize=.5\hsize}X}

\usepackage{arydshln}
\usepackage{booktabs}
\usepackage{multirow}

\usepackage{pifont}  %
\usepackage{makecell}

\newcommand{\optparens}[1]{\if\relax\detokenize{#1}\relax\else(#1)\fi}   %

\usepackage[noabbrev,capitalize]{cleveref}
\crefname{equation}{equation}{equations}
\crefname{section}{section}{sections}
\crefname{footnote}{footnote}{footnotes}   
\crefname{line}{line}{lines}   
\crefname{assumption}{assumption}{assumptions}
\crefname{lstlisting}{listing}{listings}
\Crefname{lstlisting}{Listing}{Listings}
\crefname{appendix}{Appendix}{Appendices}
\Crefname{appendix}{Appendix}{Appendices}
\usepackage{bbm}
\usepackage{appendix}

\usepackage{xcolor}  %
\usepackage{soul}
\definecolor{aigold}{RGB}{244,210, 1} 
\definecolor{aigreen}{RGB}{245, 255, 249}

\definecolor{humanpurple}{RGB}{235, 222, 240} 

\definecolor{commentgray}{RGB}{86, 101, 115}

\definecolor{light-blue}{rgb}{0.6,0.6,1}

\definecolor{aired}{RGB}{255,180,181}

\usepackage{listings}
\usepackage{parcolumns}
\lstdefinestyle{datalogstyle}{
	basicstyle={\codefont\small},  
	xleftmargin={6pt},
        xrightmargin={6pt},
        breakindent=0pt,
	frame=tb,
	stepnumber=1,
	firstnumber=1,
	numberfirstline=true,
	tabsize=2,
	showtabs=false,
	showspaces=false,
	showstringspaces=false,
	extendedchars=true,
	breaklines=true,
	columns=fullflexible,
	keepspaces=true,
	escapeinside={@}{@},
	firstnumber=last,
	captionpos=b,
	commentstyle=\color{black!65},
	numberstyle=\tiny\color{black!65},
	stringstyle=\color{codepurple},
	breakatwhitespace=false, 
	keepspaces=true,                 
	numbersep=5pt,                  
	showspaces=false,                
	showstringspaces=false,
	showtabs=false,
	aboveskip={0.8\baselineskip},
	belowskip={0.2\baselineskip},
	backgroundcolor=\color{aigreen},
}
\definecolor{rebuttal}{RGB}{229,255,204}
\sethlcolor{rebuttal}

\newcommand{\codefont}{\fontfamily{lmtt}\selectfont}

\usepackage[textwidth=2.7cm]{todonotes}

\newcommand{\cutforspace}[1]{}

\newcommand\myshade{85}
\colorlet{myurlcolor}{blue}
\hypersetup{
  citecolor  = black,
  urlcolor   = myurlcolor!\myshade!black,
  colorlinks = true,
}

\title{LookBench: A Live and Holistic Open Benchmark for Fashion Image Retrieval}

\vspace{-2cm}
\author{%
  Chao Gao$^*$, Siqiao Xue$^*$, Jiwen Fu \\[0.3em]
  \textbf{Tingyi Gu, Shanshan Li, Fan Zhou} \\[0.3em]
  Gensmo.ai \\[0.6em]   
  \textbf{\href{https://serendipityoneinc.github.io/look-bench-page/}
     {\textcolor{blue!60!black}{\faGlobe\enspace{Website}}}
    \quad
  \href{https://github.com/SerendipityOneInc/look-bench}
     {\textcolor{blue!60!black}{\faGithub\enspace{Code}}}
     \quad
    \href{https://huggingface.co/datasets/srpone/look-bench}%
      {\textcolor{blue!60!black}{\faDatabase\enspace Data}}
  \quad
  \href{https://huggingface.co/srpone/gr-lite}%
      {\textcolor{blue!60!black}{\faCogs\enspace{Model}}}
      }%
}

\begin{document}

\def\thefootnote{*}\footnotetext{These authors contributed equally to this work.}
\maketitle

\begin{abstract}


In this paper, we present \textsc{LookBench}\footnote{We use the term ``look'' to reflect retrieval that mirrors how people shop---finding the exact item, a close substitute, or a visually consistent alternative.}, a live, holistic and challenging benchmark for fashion image retrieval in real e-commerce settings. \textsc{LookBench} includes both recent product images sourced from live websites and AI-generated fashion images, reflecting contemporary trends and use cases. Each test sample is time-stamped and we intend to update the benchmark periodically, enabling contamination-aware evaluation aligned with declared training cutoffs. Grounded in our fine-grained attribute taxonomy, \textsc{LookBench} covers single-item and outfit-level retrieval across. Our experiments reveal
that \textsc{LookBench} poses a significant challenge on strong baselines, with many models achieving
below $60\%$ Recall@1. Our proprietary model achieves the best performance on \textsc{LookBench}, and we release an open-source counterpart that ranks second, with both models attaining state-of-the-art results on legacy Fashion200K evaluations. \textsc{LookBench} is designed to be updated semi-annually with new test samples and progressively harder task variants, providing a durable measure of progress. We publicly release our leaderboard, dataset, evaluation code, and trained models. 

\end{abstract}


\vspace{-0.1cm}
\begin{figure}[h]
    \centering
    \includegraphics[width=0.95\linewidth]{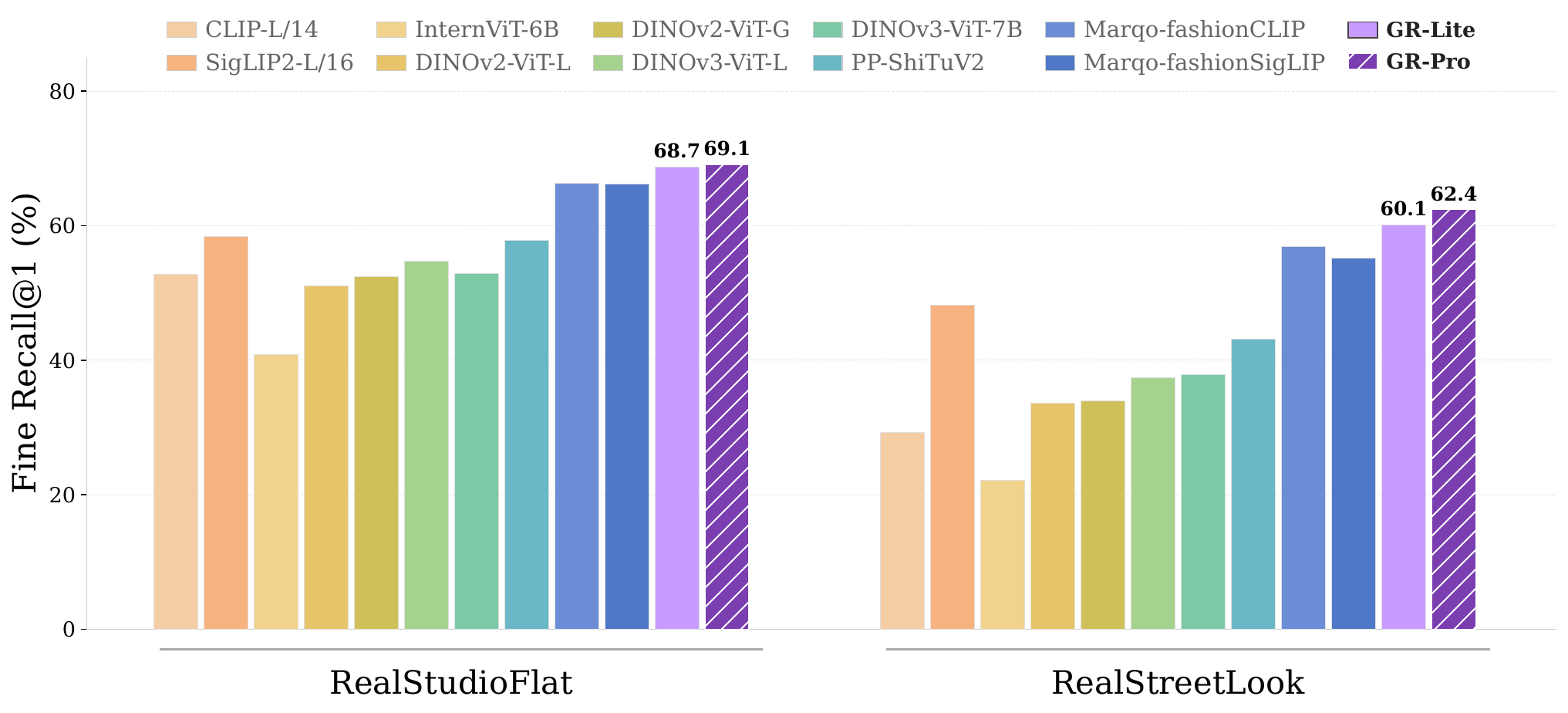}
    \caption{Fine Recall@1 of all evaluated models on the two \textsc{LookBench} subsets, RealStudioFlat and RealStreetLook. Our GR-Pro and GR-Lite models achieve the highest scores on both tasks.}
    \label{fig:placeholder}
\end{figure}

\section{Introduction}
\label{section:intro}

Fashion image retrieval has become a core capability for e-commerce platforms, powering visual search~\citep{song2015deepmetriclearninglifted}, product recommendation~\citep{yuki2025}, and outfit generation~\citep{Xu_2024}, and has inspired a large body of work on fashion-specific retrieval models~\citep{li2021embeddingtaobao,shoib2023methodsadvancement}. In contrast to these rapid modeling advances, evaluation has remained relatively stagnant: the dominant fashion benchmarks~\citep{street2shop2015,liuLQWTcvpr16DeepFashion,han2017automatic,DeepFashion2} are largely static and product–centric. Even when street-look images are included, evaluation is typically framed as instance-level, single-item retrieval on a fixed test set, making these benchmarks ill-suited for studying real user intents or for up-to-date assessment. Meanwhile, vision models such as CLIP~\citep{radford2021learningtransferablevisualmodels} and the DINO family~\citep{caron2021emerging,simeoni2025dinov3} are trained on web-crawled corpora that likely overlap with these legacy datasets, and recent vision-language models (VLMs)~\citep{Qwen-VL, Qwen2VL, gunther2025jinaembeddingsv4} reuse these backbones, making existing benchmarks vulnerable to data contamination and overfitting. Therefore, fair assessment of fashion image retrieval requires a fresh, contamination-limited protocol~\citep{jain2024livecodebench,xue2024famma}.

\begin{figure}
    \centering
    \includegraphics[width=\linewidth]{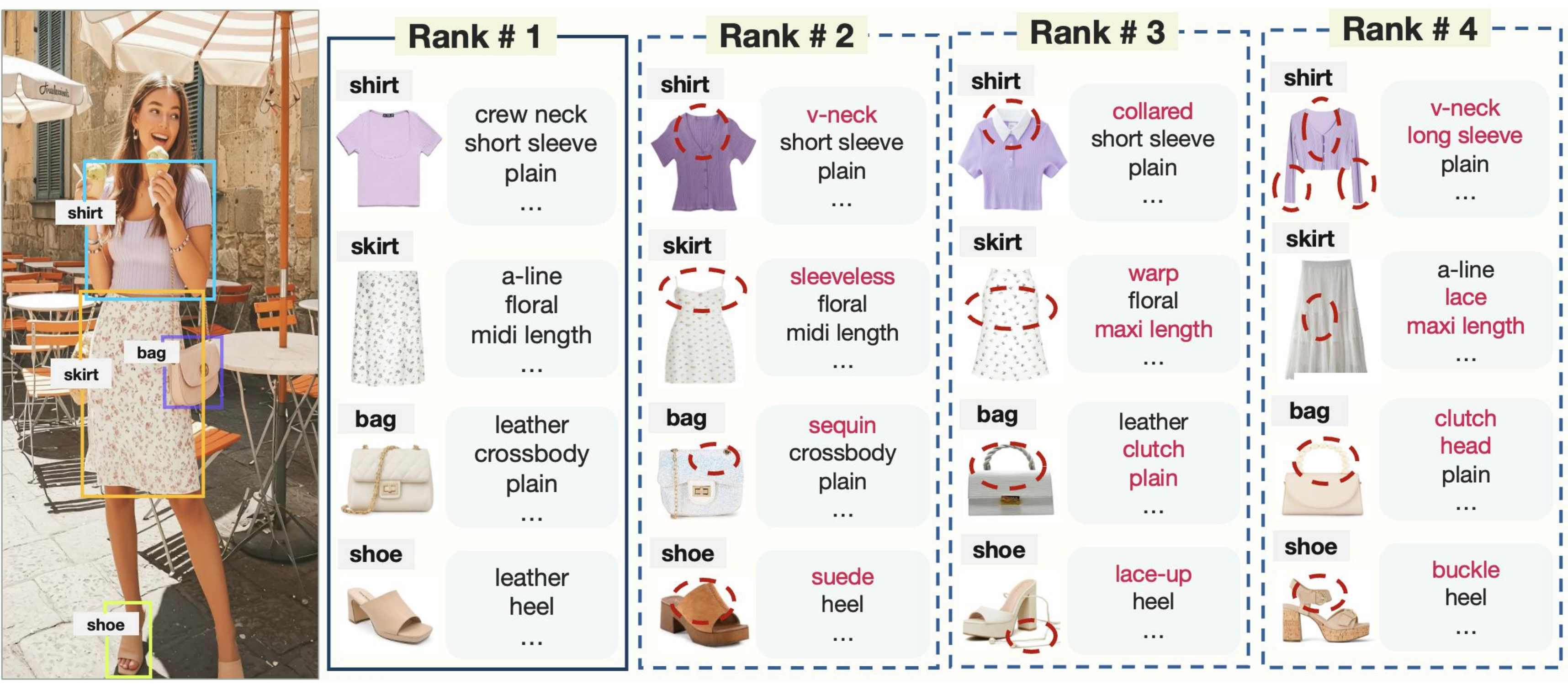}
    \caption{Illustration of the StreetLook multi-item retrieval task in \textsc{LookBench}. 
Given a street-look outfit image (left), we detect the shirt, skirt, bag and shoes and 
retrieve ranked outfit candidates from the catalog. For every retrieved product image we attach its pre-annotated attributes, and the rank (Rank \#1– \#5 in the benchmark; only Rank \#1–\#4 are shown here for readability) is determined by how many attributes match the corresponding query item. Attributes shown in red indicate attribute-level mismatches. For simplicity, this example does not distinguish between the main and other attributes; all attributes are displayed uniformly.}
    \label{fig:retro_main}
\end{figure}

In this work, we introduce \textsc{LookBench}, a live, holistic, and challenging benchmark for fashion image retrieval in real e-commerce settings. \textsc{LookBench} is built on the following principles.

\begin{itemize}[leftmargin=*]
    \item Continuously refreshing samples to mitigate contamination. 
We approximate realistic traffic via time-constrained web queries on recent retail sites, collecting studio product and street-look images, and product candidates. For each sample we log the crawl time as a proxy timestamp, which---while imperfect---enables contamination-aware evaluation by filtering test data to post–training-cutoff crawls. We periodically refresh \textsc{LookBench} with newly collected queries while preserving earlier snapshots, reducing overfitting to a single static test set. See \cref{app:query_collect} for the details of the collection process.
\item Covering diverse retrieval intents and difficulty levels. \textsc{LookBench} comprises four evaluation subsets: RealStudioFlat, AIGen-Studio, RealStreetLook and AIGen-StreetLook. RealStudioFlat contains clean flat-lay product images from real studio shoots and defines a single-item retrieval task under an ``easy'' setting. AIGen-Studio includes AI-generated product images rendered in lifestyle studio contexts, supporting single-item retrieval under a ``medium'' difficulty level. AIGen-StreetLook consists of AI-generated images depicting full street-style outfits in complex, in-the-wild scenes, supporting multi-item retrieval and representing the ``hard'' setting. Finally, RealStreetLook features real-world street-style fashion photography of complete outfits, posing the greatest visual and domain challenge, and serves as the difficult subset as well. See \cref{fig:retro_main} for an illustrative example.
The current release comprises approximately $2{,}500$ queries, each evaluated against a carefully curated retrieval corpus of about $60{,}000$ images per task; see \cref{tab:data_source} for an overview. Details of the benchmark construction are provided in \cref{section:bench}.

\item Attribute-supervised and holistic evaluation.
We construct a category-specific fashion attribute taxonomy with over 100 visually grounded properties (typically 10–25 attributes per category; see \cref{tab:attr_taxonomy} in \cref{app:annotation}) and pre-annotate all query and ranked item crops using Qwen2.5-VL-72B~\citep{Qwen2VL}. An LLM-as-a-judge audit on 200 randomly sampled crops with GPT-5.1 estimates a per-attribute correctness of roughly $93\%$, providing reliable weak supervision. These labels enable region-aware evaluation and attribute-conditioned analyses of retrieval behavior (e.g., matching materials, silhouettes, and colors), supporting more fine-grained diagnosis than category-only metrics.

\end{itemize}

Building on this taxonomy, we first benchmark a suite of strong vision--language and vision-only retrieval models. Our analysis reveals that: \circone generic open VLM backbones such as CLIP and SigLIP2~\citep{tschannen2025siglip2}, despite their strength on web-scale benchmarks, underperform on \textsc{LookBench}---particularly on the RealStreetLook subset with street-style, multi-item outfits; \circtwo fashion-specific fine-tuning consistently narrows this gap, but still leaves substantial headroom on the hardest settings. 


Motivated by these gaps, we train a family of lightweight, \emph{text-free} vision encoders on a large in-house fashion corpus under our attribute supervision. Our best proprietary encoder, \textbf{GR-Pro}, attains top-ranked performance on both legacy fashion retrieval benchmarks (e.g., Fashion200K) and \textsc{LookBench}, and a closely performing variant is released as an open model, \textbf{GR-Lite}, which provides the strongest publicly available reference on \textsc{LookBench}. Together with the open-sourced benchmark splits, attribute annotations, and evaluation code, this setup enables transparent, contamination-aware comparison and reproducible research in fashion image retrieval.

\begin{table*}[tb]
  \centering
  \small
  \renewcommand{\arraystretch}{1.2}
  \setlength{\tabcolsep}{10pt}
  \begin{sc}
  \begin{tabularx}{\textwidth}{l X C C C}
    \toprule
    Dataset & Image source & \# Retro Items & Difficulty & \# Queries/Corpus  \\
    \midrule
    RealStudioFlat       & \makecell[l]{Real studio\\flat-lay \\ product photos} & Single & Easy   & 1,011/62,226 \\
    AIGen-Studio         & \makecell[l]{AI-generated \\lifestyle\\studio images}     & Single & Medium & 192/59,254 \\
    RealStreetLook       & \makecell[l]{Real street \\ outfit photos}  & Multi  & Hard  & 1,000/61,553 \\
        AIGen-StreetLook     & \makecell[l]{AI-generated \\ street outfit \\compositions} & Multi  & Hard   & 160/58,846 \\
    \bottomrule
  \end{tabularx}
  \end{sc}
  \caption{Overview of evaluation sets in \textsc{LookBench}.}
  \label{tab:data_source}
\end{table*}

\section{Related Work}
\label{sec:related_work}

\paragraph{Vision foundation models.}
Large-scale web pretraining has produced powerful vision and vision–language encoders for retrieval. Contrastive image–text models, e.g., CLIP, ALIGN~\citep{jia2021}, learn a shared embedding space and transfer well to zero-shot tasks; broader multimodal pretraining further supports captioning and generation~\citep{li2022blip,gunther2025jinaembeddingsv4}. In parallel, vision-only self-supervised methods such as the DINO family~\citep{caron2021emerging} yield strong general visual features without text supervision and are competitive backbones for fine-grained retrieval. However, recent evidence suggests that the text encoders in contrastive VLMs can bottleneck compositional reasoning~\citep{kamath2023textencoders}, which is critical for attribute and outfit-level fashion search. In our benchmark, we report results for both families and release a strong open baseline fine-tuned on DINOv3 under our attribute taxonomy, providing a strong, text-free baseline for compositional fashion retrieval.

\paragraph{Image retrieval and benchmarks in fashion.}
Image retrieval is commonly approached via deep metric learning~\citep{wang2023introspective}, where the goal is to learn embeddings that preserve semantic similarity. Classic objectives include contrastive~\citep{chen2020big}, triple~\citep{Zhang_2024}, and their extensions~\citep{ZENG2020103820} while recent studies also question evaluation practices and sampling design~\citep{musgrave2020}. Together, these works establish the backbone for single-item and outfit-level (set) retrieval in vision. Early fashion datasets target single-item matching under controlled settings. Street2Shop~\citep{street2shop2015} frames cross-domain matching from in-the-wild street photos to shop images. DeepFashion~\citep{liuLQWTcvpr16DeepFashion} introduces in-shop and consumer-to-shop retrieval, while DeepFashion2~\citep{DeepFashion2} extends to detection, pose, segmentation, and re-identification with large consumer–commercial pairs. Fashion200K~\citep{han2017automatic} collects around 200K commercial product images with fine-grained attribute annotations and text-based attribute edits, enabling attribute-conditioned and compatibility-aware fashion retrieval. While these resources catalyzed progress, they are static and predominantly product-centric, and their images likely overlap with web-scale pretraining corpora, complicating fair evaluation at the current stage. Our benchmark continuously refreshes its test split and time-stamps samples to mitigate data contamination, while evaluating both single-item and outfit-level retrieval in real e-commerce conditions. This design broadens coverage of shopper intents and provides a durable, auditable measure of progress.

\begin{figure}
    \centering
    \includegraphics[width=0.98\linewidth]{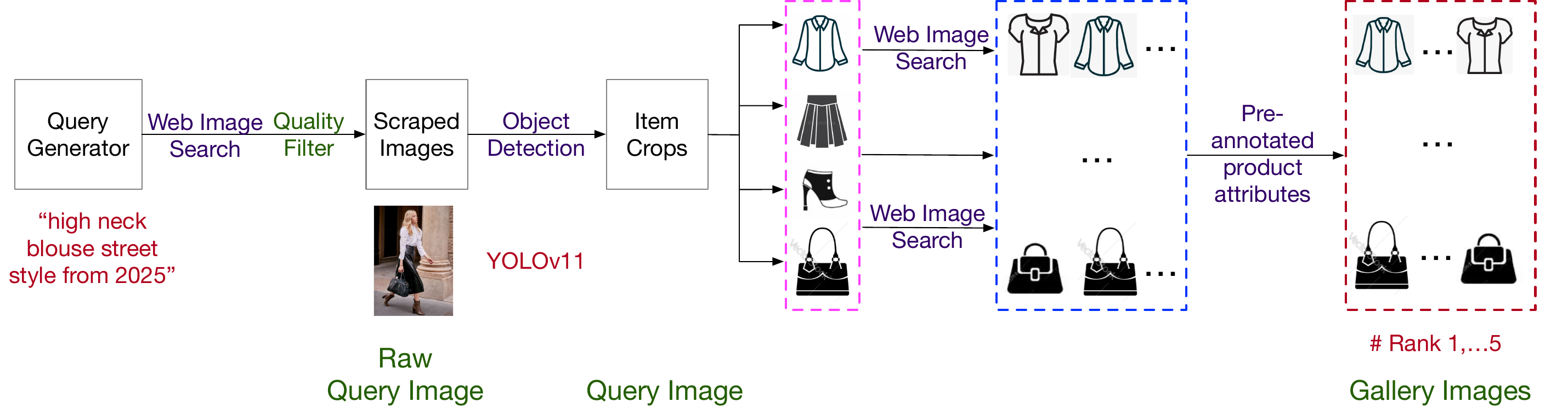}
    \caption{RealStreetLook retrieval sets construction pipeline in \textsc{LookBench}.}
    \label{fig:data_collect}
\end{figure}

\begin{figure}[t]
    \centering
    \begin{subfigure}[t]{0.45\textwidth}
        \centering
        \includegraphics[width=\linewidth]{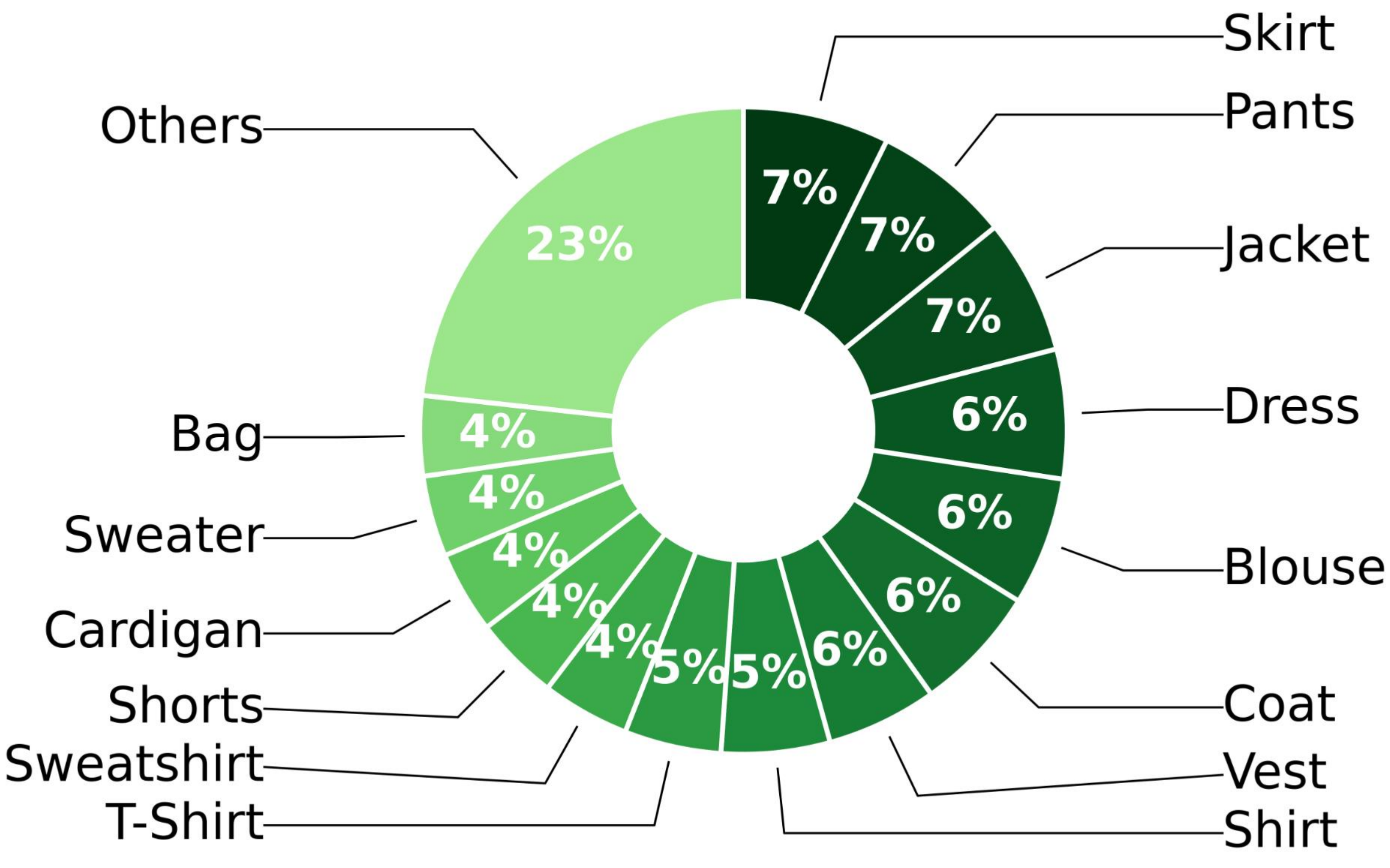}
        \caption{Category distribution of query garments across all \textsc{LookBench} tasks, spanning 27 categories including tops, outerwear, bottoms, dresses, accessories, and others.}
        \label{fig:data_statistics}
    \end{subfigure}
    \hfill
    \begin{subfigure}[t]{0.48\textwidth}
        \centering
        \includegraphics[width=\linewidth]{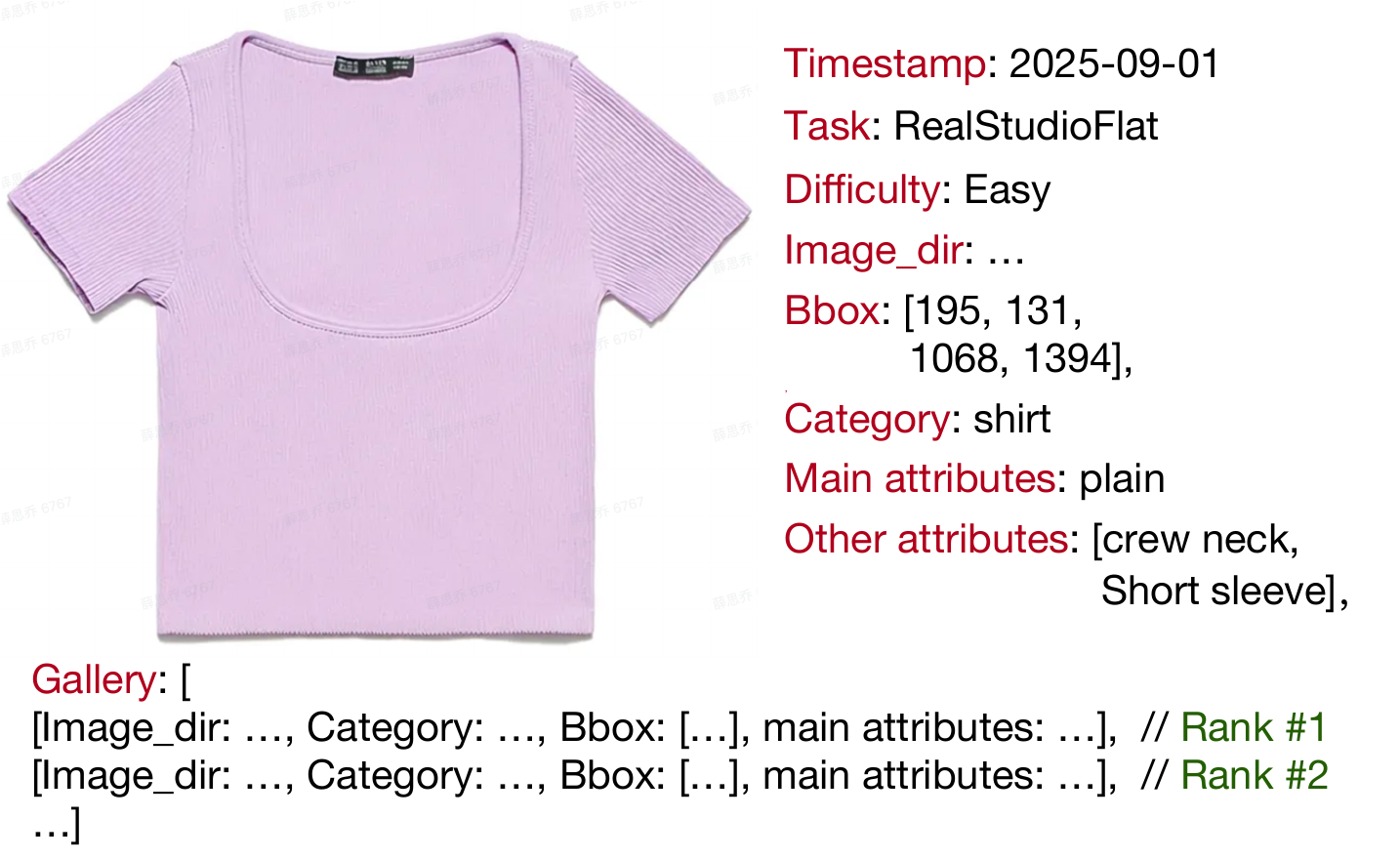}
        \caption{Example query from the \textsc{LookBench} RealStudioFlat split, 
        showing the query image and associated metadata (timestamp, task, 
        difficulty, garment category, attributes, and ranked gallery items) 
        used for retrieval evaluation.}
        \label{fig:shirt_example}
    \end{subfigure}

    \caption{\textsc{LookBench} dataset overview. 
    (a) The benchmark spans a diverse set of apparel categories. 
    (b) Each query is paired with rich metadata to support fine-grained 
    fashion image retrieval.}
    \label{fig:data_overview}
\end{figure}

\section{\textsc{LookBench} Description}
\label{section:bench}

\subsection{Benchmark Construction}

\paragraph{Query data collection.}
\textsc{LookBench} comprises four retrieval subsets---RealStudioFlat, RealStreetLook, AIGen-Studio, and AIGen-StreetLook---constructed using a unified, category--attribute–driven pipeline. Starting from our taxonomy, we sample (category, attribute, year) tuples and instantiate structured templates to form web image search queries (see \Cref{lst:query_studio_flat,lst:query_street_look}) or image-generation prompts (see \Cref{lst:prompt_aigen_studio,lst:prompt_aigen_streetlook}). 

For the real-image subsets, RealStudioFlat and RealStreetLook, these queries are issued to a commercial image search engine to retrieve time-stamped studio packshots and street-style photos, which are then de-duplicated and filtered for resolution, watermarking, and obvious off-topic noise. We apply YOLOv11~\citep{khanam2024yolov11} to each retained image to localize fashion items and obtain category-labeled crops; these crops serve as visual queries and are further enriched with fine-grained attributes using the pre-annotation pipeline described below. For each query crop, we then construct a candidate \emph{gallery}\footnote{We use the terms ``gallery'' and ``ranked results'' interchangeably throughout the paper.} by retaining only images that share the same category and main attribute, and rank these candidates by the number of additional attributes they share with the query, keeping the top-$k$ ranked results as positives. An illustrative collection pipeline of RealStreetLook data is show in \cref{fig:data_collect}. The AIGen-Studio and AIGen-StreetLook subsets follow the same detection, attribute labeling, and ranking procedure but differ in how the base imagery is obtained: for each (category, attribute) pair, we sample scene variables such as location, lighting, camera angle, and layout from curated placeholder lists, feed them to Qwen2.5-VL-72B to generate detailed textual descriptions, and then condition Qwen-Image~\citep{wu2025qwenimagetechnicalreport} to synthesize high-resolution studio or street-look images. Further implementation details are provided in \cref{app:query_collect}.

\paragraph{Corpus data collection.}
For each evaluation subset, we construct its retrieval corpus by augmenting the ranked gallery images with task-specific distractors from Fashion200K. We embed all \textsc{LookBench} queries and gallery images, as well as all Fashion200K images, using CLIP ViT-L/14, and for each task treat the union of its queries and galleries as anchor points in this embedding space. A Fashion200K image is kept as a distractor for a given task only if its cosine similarity to at least one anchor falls within a task-specific band $[\alpha_t, \beta_t]$, which removes near-duplicates ($>\beta_t$) and clearly unrelated items ($<\alpha_t$) while preserving visually plausible but non-matching ``soft negatives''. A detailed description of the procedure used to select $[\alpha_t, \beta_t]$ for each subset, together with the resulting thresholds and numbers of distractors summarized in \cref{tab:corpus_noise}, is provided in \cref{app:query_collect}.

After augmenting with these distractors, the corpora contain 62{,}226 images for RealStudioFlat, 59{,}254 for AIGen-Studio, 61{,}553 for RealStreetLook and 58{,}846 for AIGen-StreetLook, supporting 1{,}011, 192, 1{,}000, and 160 queries respectively (see \cref{tab:data_source}). This construction yields four parallel, contamination-aware retrieval benchmarks with aligned category-attribute structure across real and synthetic domains.

\paragraph{Pre-annotation.}
To equip all four \textsc{LookBench} evaluation sets with fine-grained supervision, we pre-annotate attributes for both query images and their associated gallery items. For each image, we first detect individual garments and extract item crops. Each crop is then fed to Qwen2.5-VL-72B together with (i) its fashion category and (ii) the corresponding category-specific attribute vocabulary from our taxonomy. The model is instructed to focus only on the target item and to select attributes that are clearly supported by visual evidence, returning a structured JSON object with a single \texttt{main\_attribute} (the most salient property) and a list of \texttt{other\_attributes} that are also visible, without introducing attributes outside the provided vocabulary. We run this pipeline offline over all query items and over the gallery items that appear in our ranked lists, so that every evaluated query–result pair in all four tasks is accompanied by aligned attribute labels. \cref{fig:data_statistics} summarizes the category distribution, while \cref{fig:shirt_example} illustrates the structure of an individual query image.

To estimate annotation quality, we then apply GPT-5.1 in an \emph{LLM-as-a-judge} protocol: given the crop, category, attribute vocabulary, and Qwen predictions, the judge model independently marks each predicted attribute (main and other) as correct or incorrect based solely on the image. On a random sample of 200 item crops, this yields an estimated attribute-level correctness of approximately 93\%, indicating that the pre-annotations provide high-precision weak supervision for downstream retrieval evaluation and attribute-level analysis. Further details of the taxonomy (\cref{tab:attr_taxonomy}), prompt templates (\Cref{lst:attr_annot_prompt}), and judging setup (\Cref{lst:attr_judge_prompt}) are provided in \cref{app:annotation}.

\subsection{Update and Maintenance Plan}
Maintaining a contamination-limited benchmark requires regularly extending the pool of evaluation queries and tasks. \textsc{LookBench} currently has one release, and the next scheduled release will:
\begin{itemize}[leftmargin=*]
    \item scale up existing tasks, expanding the benchmark to roughly 10K total queries and a corpus of about 200K images, while preserving the current split definitions.
    \item add new, harder retrieval subtasks, such as more challenging cross-domain and outfit-level settings, built on top of this expanded core.
\end{itemize}

\section{Our Model: GensmoRetro}
\label{sec:models}

Beyond serving as an evaluation benchmark, \textsc{LookBench} also motivates the design of strong baselines that are practical for real e-commerce use. In this section we describe our family of \emph{text-free} visual encoders, GensmoRetro (GR).

\begin{itemize}[leftmargin=*]
    \item \textbf{GR-Pro} is a high-capacity, single-tower vision encoder built on a
    Vision Transformer (ViT) backbone with approximately 0.3B trainable parameters. The network follows the standard design of
    a deep feature extractor with global pooling, producing an $\ell_2$-normalized
    image embedding for retrieval. We train GR-Pro on our in-house fashion corpus
    of 6.5\,M images (see \cref{tab:data_for_gr_model}). For commercial reasons,
    we do not disclose its exact architectural configuration or release its weights, but GR-Pro is accessible via an API under
    a separate usage license upon request.

   \item \textbf{GR-Lite} is a separate lightweight vision encoder that we design
for public release. It uses a DINOv3 backbone with a parameter count comparable
to that of GR-Pro, followed by a linear projection to a $d$-dimensional
retrieval embedding. GR-Lite is trained on a smaller fashion corpus consisting of 1.3\,M open-source images and 0.5\,M in-house images. Unlike
    GR-Pro, we fully disclose GR-Lite's architecture and training
    method in \cref{sec:model_arch} and \cref{sec:training}, respectively, and release its pretrained weights, providing a strong, fully
    reproducible reference model on \textsc{LookBench}.
\end{itemize}

\subsection{Model Architecture of GR-Lite}
\label{sec:model_arch}
GR-Lite is instantiated as a DINOv3 ViT-L/16 backbone followed by a lightweight trainable projection head. Concretely, we start from the public checkpoint\footnote{\small\url{https://huggingface.co/facebook/dinov3-vitl16-pretrain-lvd1689m}}, a 24-layer ViT-L/16 with $16{\small\times}16$ patches and roughly 300M parameters, pretrained in a self-supervised manner on a large-scale curated web image corpus. The backbone produces a 1024-dimensional global representation for each image, which we map to a 512-dimensional embedding and then $\ell_2$-normalize; this 512-dimensional vector is used as the final retrieval representation at inference time.

This architecture is well suited to fashion image retrieval. The backbone provides high-capacity, patch-level features that capture fine-grained cues such as textures, patterns, and accessories, which are crucial for instance-level retrieval~\citep{caron2021emerging}. Web-scale self-supervised pretraining yields strong, transferable representations~\citep{radford2021learningtransferablevisualmodels}, enabling GR to adapt to real-world e-commerce imagery in a purely visual setting. At the same time, the backbone strikes a favorable balance between accuracy and inference-time efficiency, making GR practical for latency and memory-constrained systems.

\subsection{Large Scale Training of GR-Lite}
\label{sec:training}
We perform large scale fully fine-tuning of both the backbone and the 512-dimensional projection head on a large in-house corpus of fashion images spanning studio product photos and in-the-wild outfits. This stage adapts the generic DINOv3 representations to the fashion domain while keeping the encoder purely visual and text-free.

\paragraph{Data collection and curation.}
Our training corpus combines crawled e-commerce imagery with several open-source fashion datasets. On the e-commerce side, we scrape product detail pages and collect all associated images (packshots, on-model views, color variants), assigning them a shared product identity label so that images from the same page form one identity group. For public datasets, we reuse the provided product or identity annotations when available and map them into the same label space. Because these labels can still be noisy, we perform a second round of curation with Qwen2.5-VL-72B, using it as a visual verifier within each identity group and discarding images that do not clearly depict the same product. This procedure yields weakly supervised but high-precision product identities at scale. \cref{tab:data_for_gr_model} summarizes the resulting image counts and identities per source.

\paragraph{Learning objective.}
Given the embeddings produced by the projection head, we train our model with an additive
angular-margin softmax loss---ArcFace loss~\citep{Deng_2022}. Let $d$ be the embedding dimension and
$f_i \in \mathbb{R}^d$ the embedding of sample $i$. The classifier is parameterized by
\(W = [W_1,\dots,W_C] \in \mathbb{R}^{d \times C}\),
where \(C\) is the number of product identities and
\(W_j \in \mathbb{R}^d\) is the class center for identity \(j\). We $\ell_2$-normalize both features and class weights so that
\[
\cos\theta_{i,j} = W_j^\top f_i, \qquad \|W_j\|_2 = \|f_i\|_2 = 1.
\]
For a sample with ground-truth identity $y_i$, the per-sample loss is
\begin{eqnarray}
\ell_i
= - \log
\frac{\exp\!\big(s\big(\cos(m_1\theta_{i,y_i} + m_2) - m_3\big)\big)}
     {\exp\!\big(s\big(\cos(m_1\theta_{i,y_i} + m_2) - m_3\big)\big)
      + \sum_{j\neq y_i} \exp\!\big(s \cos\theta_{i,j}\big)} ,
\label{eq:arcface}
\end{eqnarray}
and the overall objective is $\mathcal{L}=\tfrac{1}{N}\sum_{i=1}^N \ell_i$.
Throughout experiments we use $m_1{=}1.0$, $m_2{=}0.25$, $m_3{=}0.0$ and a scale factor
$s{=}32$, which corresponds to the standard ArcFace configuration.
Minimizing~\Cref{eq:arcface} increases the target logit
$\cos(m_1\theta_{i,y_i}+m_2)-m_3$ and decreases $\cos\theta_{i,j}$ for $j\neq y_i$, i.e.,
it drives $\theta_{i,y_i}$ towards $0$ while enlarging the angles to other class centers.
Since all images of the same product share the same identity $y$ and therefore the same
center $W_y$, their embeddings are jointly pulled toward $W_y$, making same-label
embeddings close to each other and far from other identities.
The logits $s\cos\theta_{i,j}$ are realized via a distributed classifier implemented
with Partial FC~\citep{an2022killingbirdsstoneefficientrobust}, which maintains all class
centers but updates only the subset activated in each iteration, yielding discriminative
embeddings at large scale.

\paragraph{Mini-batch construction.}
At each training step we form a mini-batch by sampling product identities and, for each
identity, sampling multiple images from its group. A typical batch therefore contains
tuples of the form $(x_1,y_1),(x_2,y_1),(x_3,y_1),(x_4,y_2),(x_5,y_3),(x_6,y_4),\ldots$,
where images sharing the same label $y$ come from the same product page or open-source
identity. This sampling scheme exposes the loss to diverse views of each product while
keeping supervision simple and identity-based.

\paragraph{Optimization.}

We follow the DINOv3 optimization recipe and train with AdamW~\citep{loshchilov2019} using constant learning rate. Unless otherwise noted, we use a base learning rate of $1\times 10^{-4}$ for the backbone and projection head, and a $10\times$ larger learning rate for the Partial~FC classifier to speed up convergence of the class centers. We use a global batch size of 2048 images (256 per GPU across 8 GPUs) and train with square inputs of size $224 \times 224$ pixels.

\paragraph{Data augmentation.}
For training, we use an AutoAugment-style policy~\citep{cubuk2019autoaugment} combined with
bicubic resizing, mild color jitter, and Random Erasing. During the early stages of optimization,
we additionally apply Mixup ($\alpha = 0.2$) and CutMix ($\alpha = 1.0$) with a switching
probability of 0.5. This augmentation scheme yields substantial appearance diversity while
remaining consistent with common practice for modern vision backbones.\footnote{Our
implementation follows the \texttt{timm} library~\citep{rw2019timm}, using the
\texttt{rand-m9-mstd0.5-inc1} AutoAugment policy and the default \texttt{reprob} setting for
Random Erasing.}

\begin{table*}[tb]
  \centering
  \small
  \renewcommand{\arraystretch}{1.1}
  \setlength{\tabcolsep}{10pt}
  \begin{sc}
  \begin{tabularx}{\textwidth}{l  C C C}
    \toprule
    Dataset &  \makecell[C]{\# \\ Images}  & \makecell[C]{\# \\ Identities} & Share (\%) \\
    \midrule
    DeepFashion            & 52{,}340    & 7{,}994    & 0.8 \\
    DeepFashion2          & 290{,}229   & 48{,}228   & 4.5 \\
    Fashion Product~\citep{kaggle_fashion_dataset}      & 43{,}560    & 1{,}871    & 0.6 \\
    Kream \& OnTheLook~\citep{kream_hf_dataset,onthelook_hf_dataset}    & 588{,}790   & 294{,}395  & 9.1 \\
    Product10K~\citep{bai2020products10k}           & 141{,}931   & 9{,}691    & 2.2 \\
    Watch and Buy~\citep{tianchi_531893}              & 272{,}219   & 56{,}837   & 4.2 \\
    In-house                & 5{,}114{,}170 & 1{,}485{,}522 & 78.6 \\
    \midrule
    Total                     & 6{,}503{,}239 & 1{,}904{,}538 & 100.0 \\
    \bottomrule
  \end{tabularx}
  \end{sc}
  \caption{Details of training data for GR models. \# Identities denotes the number of unique product labels after cleaning.}
  \label{tab:data_for_gr_model}
\end{table*}

\begin{table*}[tb]
  \centering
  \small
  \renewcommand{\arraystretch}{1.12}
  \setlength{\tabcolsep}{3.5pt}
  \begin{sc}
  \begin{tabularx}{\textwidth}{l  *{1}{l}*{4}{S}*{1}{S}}
    \toprule
    Model
      & Resolution
      & \multicolumn{2}{c}{AIgen}
      & \multicolumn{2}{c}{Real}
      & Overall \\
      \cmidrule(lr){3-4} \cmidrule(lr){5-6}
      & / Emb. Size & StreetLook & Studio & StreetLook & Studio &  \\
    \midrule
    \rowcolor{aigreen}%
    GR-Pro          & 336 / 1024 & \textbf{77.50} & \textbf{75.13} & \textbf{62.39} & \textbf{69.14} & \textbf{67.38} \\
    \rowcolor{aigreen!70}%
    GR-Lite         & 336 / 1024 & 75.00 & 70.47 & 60.14 & 68.74 & 65.71 \\
    Marqo-fashionCLIP    & 224 / 512  & 71.88 & 72.02 & 56.88 & 66.37 & 63.24 \\
    Marqo-fashionSigLIP  & 224 / 768  & 74.38 & 74.09 & 55.15 & 66.17 & 62.77 \\
    SigLIP2-B/16         & 384 / 768  & 69.38 & 72.02 & 51.89 & 62.81 & 59.44 \\
    SigLIP2-L/16         & 384 / 1024 & 60.00 & 65.80 & 48.22 & 58.36 & 54.84 \\
    PP-ShiTuV2           & 224 / 512  & 38.12 & 44.04 & 43.12 & 57.86 & 49.21 \\
    DINOv3-ViT-L         & 224 / 1024 & 22.50 & 38.86 & 37.41 & 54.70 & 43.97 \\
    DINOv3-ViT-7B        & 224 / 4096 & 20.00 & 39.90 & 37.92 & 52.92 & 43.33 \\
    DINOv2-ViT-L         & 224 / 1024 & 29.38 & 36.27 & 33.64 & 51.04 & 41.07 \\
    DINOv2-ViT-G         & 224 / 1536 & 38.50 & 35.75 & 33.94 & 52.42 & 41.49 \\
    CLIP-L/14            & 336 / 768  & 30.00 & 33.16 & 29.26 & 52.82 & 39.79 \\
    DINOv3-ConvNext      & 224 / 1536 & 12.50 & 25.39 & 25.08 & 45.10 & 32.88 \\
    CLIP-B/16            & 224 / 512  & 22.50 & 17.62 & 22.43 & 45.30 & 31.90 \\
    InternViT-6B         & 448 / 1024 & 19.38 & 29.53 & 22.12 & 40.85 & 30.62 \\
    \bottomrule
  \end{tabularx}
  \end{sc}
  \caption{
    Comparison of \textsc{fine-Recall@1} across the four \textsc{LookBench} subsets.
    ``Overall'' is a query-count-weighted average using dataset sizes.
    Best values per column are shown in \textbf{bold}.
  }
  \label{tab:finerecall1_clean}
\end{table*}

\subsection{Experimental Setup}

\paragraph{Benchmarked models.} We evaluate a diverse suite of state-of-the-art models: 
\begin{itemize}[leftmargin=*]
    \item Generic VLM backbones: CLIP ViT-B/16 and CLIP ViT-L/14, and SigLIP2-B, SigLIP2-L, and SigLIP2-So400m, adopted as strong generic baselines for image retrieval~\citep{song2025survey}.
    \item Fine-tuned VLM backbones: Marqo-FashionCLIP and FashionSigLIP~\citep{zhu2024generalized}, which are further trained on large-scale e-commerce data and currently report leading performance on fashion retrieval benchmarks such as DeepFashion.
    \item Vision-only models: PP-ShiTuV2~\citep{wei2022ppshitupracticallightweightimage}, InternViT-6B~\citep{chen2024expanding}, and the DINO family, including DINOv2 ViT-L/16 and ViT-G/14, and 
DINOv3 ViT-L/16, ViT-7B/16, and ConvNeXt-L~\citep{liu2022convnet} \footnote{CNN backbone trained with the DINOv3 objective}, which are commonly used in large-scale image retrieval and recognition systems.
    \item Our own vision encoders: GR-Pro and GR-Lite.
\end{itemize}

\paragraph{Metrics.}
Based on standard retrieval practice and our category-attribute taxonomy, we adopt an attribute-aware evaluation protocol that focuses on fine-grained correctness of the top-ranked result. We report the following three metrics:
\begin{itemize}[leftmargin=*]
  \item \textbf{Fine Recall@1.}
  Our primary headline metric is \emph{fine} Recall@1, which counts a query as correct only if the top-ranked item matches both the query’s garment category and all of its annotated attributes. This provides a strict measure of attribute-level retrieval fidelity and reflects whether the system surfaces a visually and semantically precise match in the position users see first.

  \item \textbf{Coarse Recall@1.}
  As a more permissive counterpart, \emph{coarse} Recall@1 credits a hit whenever the top-ranked item matches the query’s garment category, regardless of attribute agreement. This metric captures whether the model at least retrieves a category-consistent item at rank~1.

  \item \textbf{nDCG@5.}
  To assess the quality of the ranked list beyond the first item, we compute nDCG@5 using graded relevance scores derived from category and attribute overlap. This rewards models that place more attribute-consistent results higher within the top-5 positions.
\end{itemize}

For completeness, we also compute standard retrieval metrics such as fine/coarse Recall@$k$ and nDCG@$k$ for larger values of $k$. See \cref{app:metrics} for the formal definitions of the metrics. Due to space constraints, unless otherwise stated, all results reported in the main text use \emph{fine} Recall@1.

\subsection{Results and Analysis}
\label{sec:analysis}
\paragraph{Main results.} \cref{tab:finerecall1_clean} reports fine Recall@1 on all four \textsc{LookBench} subsets. Our GR-Pro model ranks first in every column, obtaining 77.50\% / 75.13\% on AIGen-StreetLook / AIGen-Studio and 62.39\% / 69.14\% on RealStreetLook / RealStudioFlat, for an overall score of 67.38\%. The open GR-Lite variant is close behind (65.71\% overall), trailing GR-Pro by only 1.67 \emph{absolute} points, while still outperforming all public baselines. The strongest competitors are Marqo-fashionCLIP and Marqo-fashionSigLIP, which are likewise fine-tuned on fashion/e-commerce data; the best of them, Marqo-fashionCLIP, reaches 63.24\% overall. GR-Pro improves on this by 4.14 \emph{absolute} points (about 6.5\% \emph{relative}).

Across subsets, GR-Pro consistently outperforms the strongest Marqo model. On the two real-image tracks, the gains are 5.5 and 2.8 \emph{absolute} points on RealStreetLook and RealStudioFlat (roughly 9.7\% and 4.2\% \emph{relative}), while on AIGen-StreetLook and AIGen-Studio the margins are 5.6 and 3.1 points. RealStreetLook is clearly the most challenging setting, with all methods dropping compared to studio photos, yet our models still maintain a clear lead. In contrast, generic encoders such as CLIP-B/L, DINOv2/3, and InternViT are substantially weaker, with overall fine Recall@1 in the 30--45\% range, indicating that strong generic vision or VLM backbones alone are not sufficient for attribute-faithful fashion retrieval.

Results for Coarse Recall@1 and nDCG@5 are reported in \cref{tab:coarse_clean} and \cref{tab:ndcg_clean} in \cref{app:exp}, and exhibit the same overall pattern, with GR-Pro and GR-Lite consistently outperforming all baselines.

\begin{figure}[htbp]
    \centering
    \begin{minipage}[t]{0.48\textwidth}
        \centering
        \includegraphics[width=\textwidth]{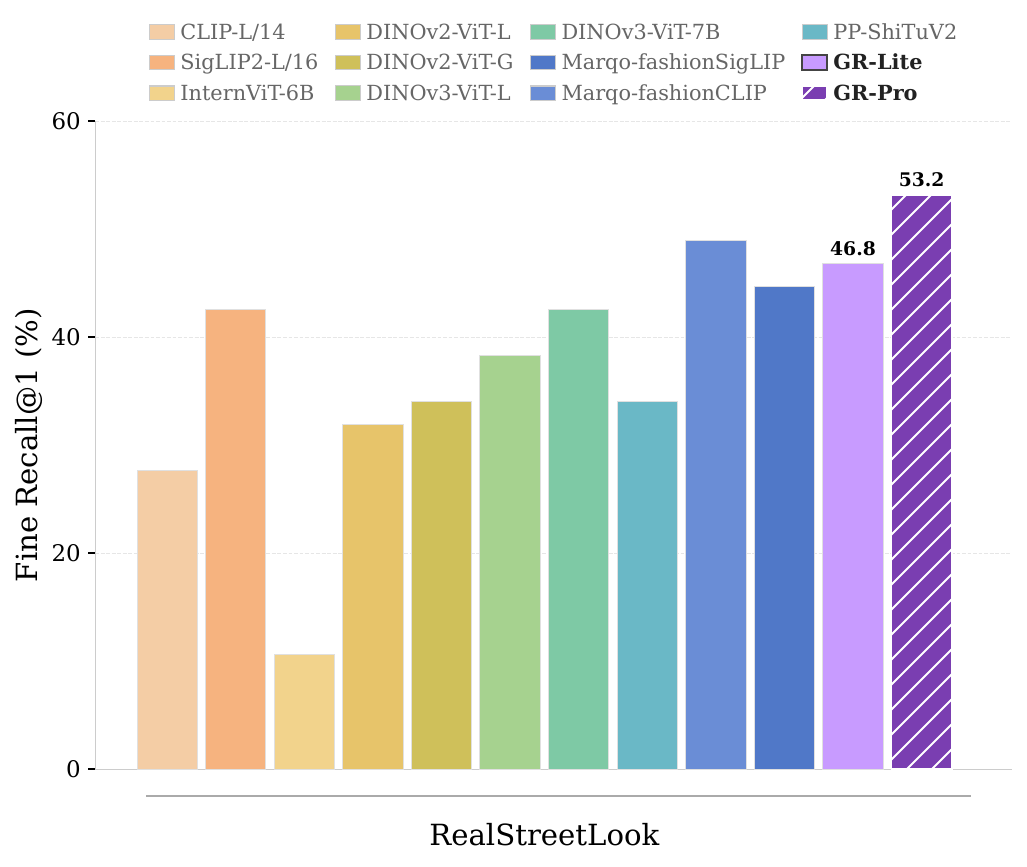}
        \caption{Outfit-level fine-grained retrieval performance on RealStreetLook subset of \textsc{LookBench}.}
        \label{fig:outfit_level}
    \end{minipage}%
    \hfill 
    \begin{minipage}[t]{0.49\textwidth}
        \centering
        \includegraphics[width=\textwidth]{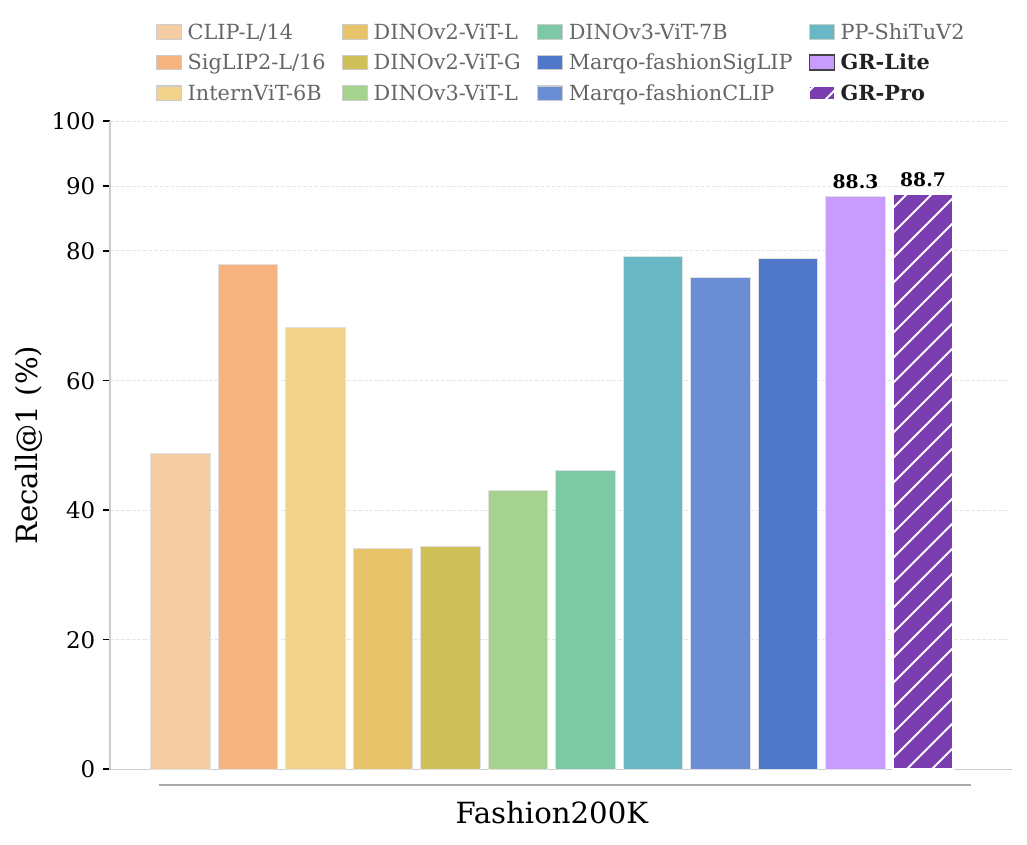}
        \caption{Single-item retrieval performance on the Fashion200K benchmark.}        
        \label{fig:fashion200k}
    \end{minipage}
    
\end{figure}

\paragraph{Analysis I: Outfit-level retrieval on \textsc{LookBench}.}
We study outfit-level retrieval on the RealStreetLook subset, where a query is counted as correct only if \emph{all} items in the outfit achieve fine Recall@1 (i.e., each garment’s top-ranked match is category- and attribute-consistent). \cref{fig:outfit_level} reports this outfit-level fine Recall@1 for all models. Generic encoders such as CLIP, DINO, InternViT and PP-ShiTuV2 achieve at most low–mid 40\% on this challenging setting, while the fashion-tuned Marqo-fashionCLIP and Marqo-fashionSigLIP push performance further. Our GR models establish a new best: GR-Lite reaches 46.8\%, and GR-Pro attains 53.2\%, clearly outperforming all baselines yet still leaving substantial headroom, highlighting the difficulty of attribute-faithful outfit retrieval.

\paragraph{Analysis II: Generalization to legacy benchmarks.}
We evaluate the same suite of models on conventional fashion retrieval benchmarks and compare their behavior to that on \textsc{LookBench} to study domain transfer. On Fashion200K, a single-image product search benchmark with catalog-style photos, generic CLIP/SigLIP encoders already perform strongly, whereas self-supervised DINO and InternViT variants lag behind. Fashion-tuned models from Marqo (Marqo-fashionCLIP and Marqo-fashionSigLIP) further boost performance, reaching around $80\%$ Recall@1. Remarkably, our GR-Lite and GR-Pro models still obtain the best results on Fashion200K ($88.3\%$ and $88.7\%$ Recall@1, respectively; see \cref{fig:fashion200k}). This indicates that the garment-aware representations induced by our attribute-supervised training on large-scale fashion data (combining open-source and in-house corpora) not only excel on the challenging \textsc{LookBench} evaluation, but also transfer robustly to standard single-item benchmarks, highlighting their broad utility for fashion retrieval.

\begin{figure}[htbp]
    \centering
    \begin{minipage}[t]{0.48\textwidth}
        \centering
        \includegraphics[width=\textwidth]{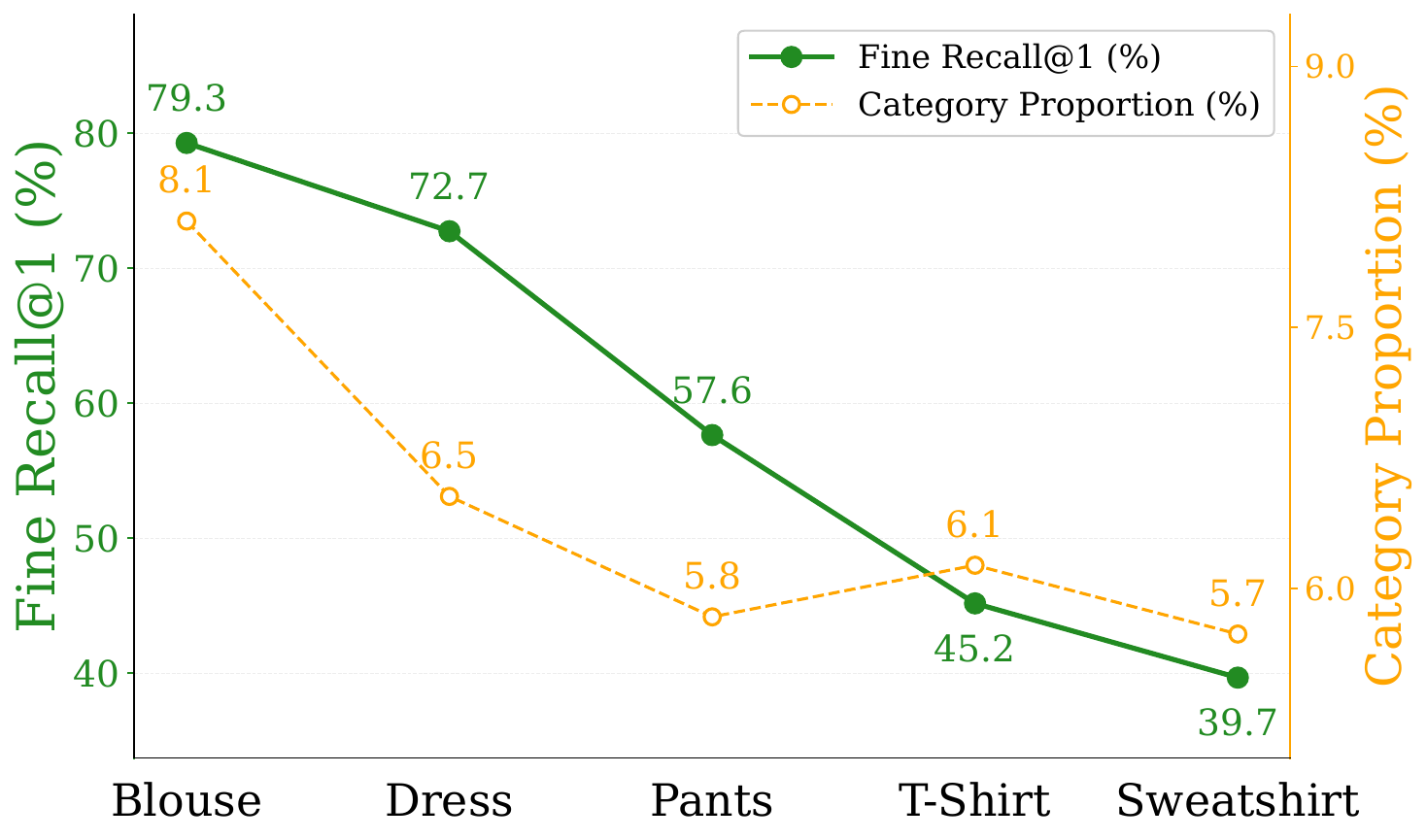}
        \caption{Fine Recall@1 of GR-Pro on the top-5 most frequent garment categories in the RealStudioFlat subset of \textsc{LookBench}.}
        \label{fig:real_studio_falt_top5_cateogory}
    \end{minipage}%
    \hfill 
    \begin{minipage}[t]{0.49\textwidth}
        \centering
        \includegraphics[width=\textwidth]{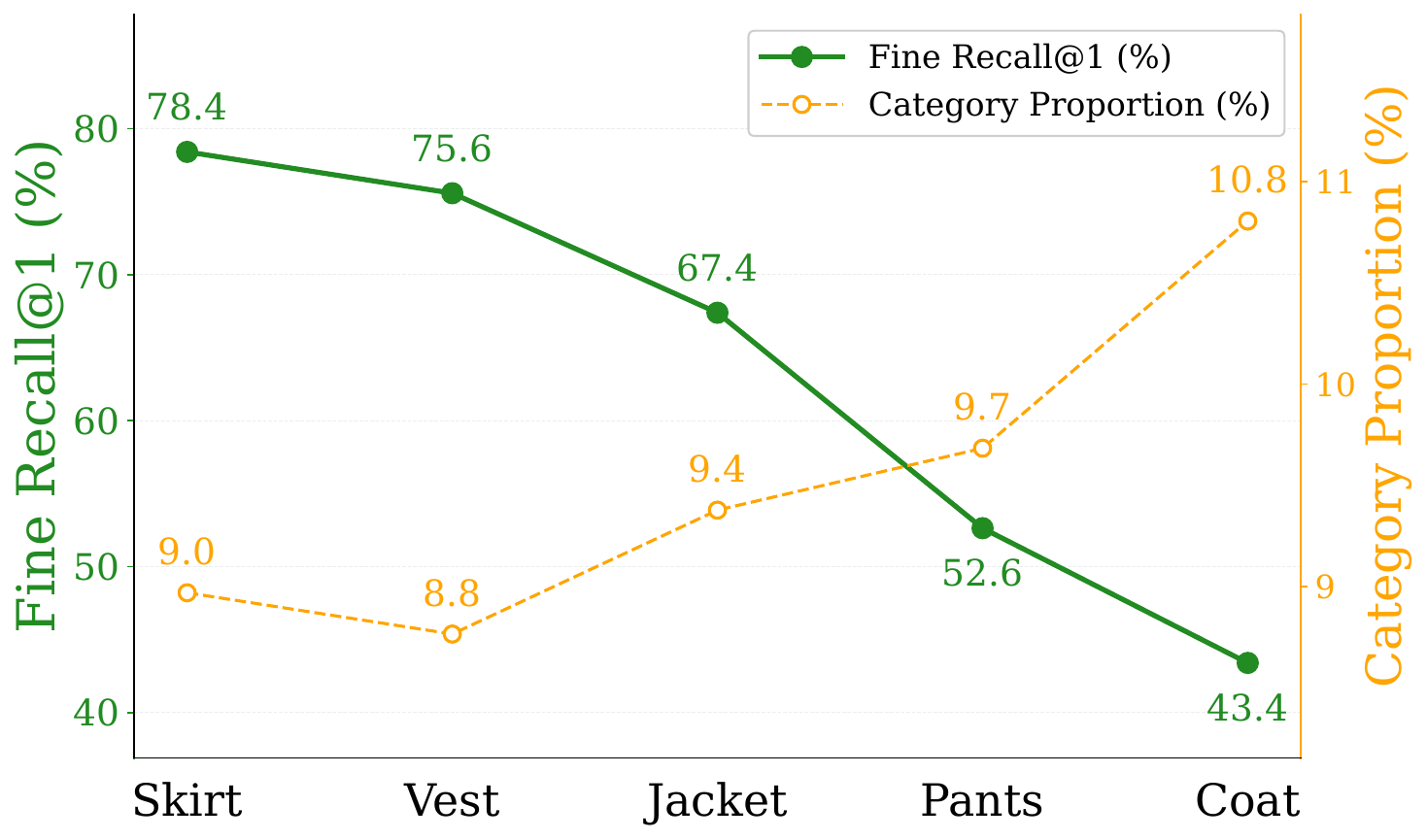}
        \caption{Fine Recall@1 of GR-Pro on the top-5 most frequent garment categories in the RealStreatLook subset of \textsc{LookBench}.}        
        \label{fig:real_streetlook_top5_cateogory}
    \end{minipage}
    
\end{figure}

\paragraph{Analysis III: Category-wise performance.}
\cref{fig:real_studio_falt_top5_cateogory} and \cref{fig:real_streetlook_top5_cateogory} break down GR-Pro’s fine Recall@1 by garment category on the RealStudioFlat and RealStreetLook subsets. On RealStudioFlat, GR-Pro is strong on visually distinctive tops and one-piece garments, reaching 79.3\% and 72.7\% Recall@1 for blouses and dresses, but performance drops to 45.2\% for T-shirts and 39.7\% for sweatshirts, even though these categories have a similar number of queries, with pants in between at 57.6\%. On RealStreetLook, where garments appear in full outfits under cluttered backgrounds and occlusions, we again observe a wide spread: skirts and vests are relatively easy (78.4\% and 75.6\%), whereas coats and pants remain challenging (43.4\% and 52.6\%). Comparing pants across domains (57.6\% in studio vs.\ 52.6\% in street) illustrates the additional difficulty of street photography. Overall, GR-Pro works best on categories with distinctive silhouettes such as dresses and skirts, but still struggles on casual or heavily layered pieces like T-shirts, sweatshirts, and coats that exhibit high intra-class similarity.

\begin{figure}[htbp]
    \centering
    \begin{minipage}[t]{0.48\textwidth}
        \centering
        \includegraphics[width=\textwidth]{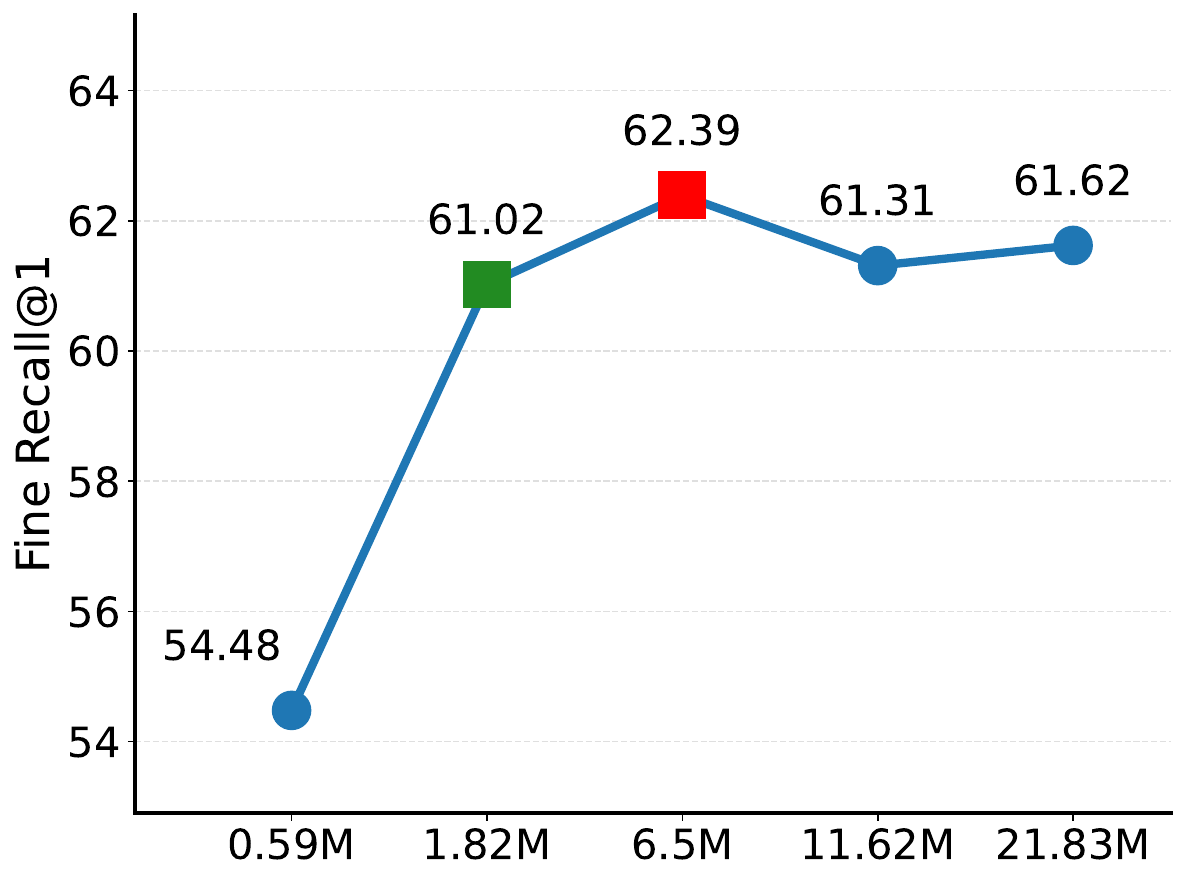}
        \caption{Data scaling on RealStreetLook subset of \textsc{LookBench}. We report fine Recall@1 with varying size of the fashion training corpus; the red square marks the 6.5 M configuration used for GR-Pro, and the green 
square marks the configuration corresponding to the open GR-Lite model.}
        \label{fig:data_scaling_law}
    \end{minipage}%
    \hfill 
    \begin{minipage}[t]{0.49\textwidth}
        \centering
        \includegraphics[width=\textwidth]{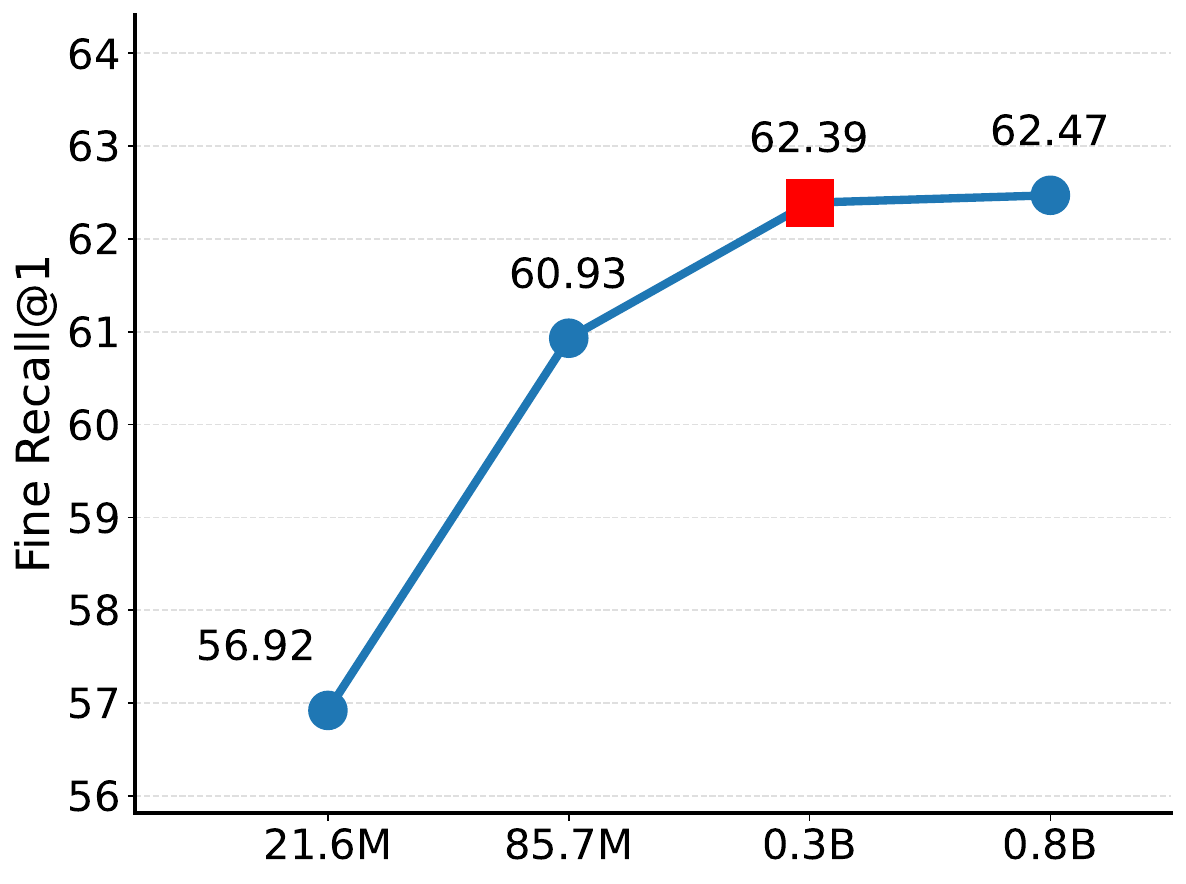}
        \caption{Model scaling on the RealStreetLook subset of \textsc{LookBench}. We report fine Recall@1 when varying the backbone size with the training corpus fixed at 6.5 M images; the red square marks the 0.3 B-parameter backbone used for GR-Pro.}        
        \label{fig:model_scaling_law}
    \end{minipage}
    
\end{figure}

\paragraph{Analysis IV: Data scaling.}
We study how retrieval performance scales with the amount of training data. Fixing the backbone architecture and all optimization hyperparameters, we vary the size of the fashion training corpus for our GR-Pro configuration and measure fine Recall@1 on the RealStreetLook subset. As shown in \cref{fig:data_scaling_law}, performance increases sharply when growing the corpus from 0.59 M to 1.82 M images (54.48\% to 61.02\%, a 6.54-point \emph{absolute} gain, or about 12\% \emph{relative}), and then improves more mildly up to 6.5 M images, where GR-Pro peaks at 62.39\%. Beyond this point, further scaling to 11.62 M and 21.83 M images yields no additional gains and even slightly degrades performance (61.31\%--61.62\%), suggesting that RealStreetLook accuracy is close to saturated for this backbone and training recipe. We therefore adopt the 6.5 M setting---corresponding to GR-Pro---as our default (see \cref{tab:data_for_gr_model}), since it lies near the knee of the data--performance curve.

\textbf{Analysis V: Model scaling.} We fix the training corpus at 6.5 M images and vary the capacity of the visual encoder, training DINOv3 backbones from 21.6 M to 0.8 B parameters with matching embedding dimensionalities. As shown in Figure~10, fine Recall@1 on RealStreetLook improves substantially when moving from the smallest to the mid-size model (56.92\% to 60.93\%, a 4.0-point absolute gain, or about 7.0\% relative), and then increases more modestly up to the 0.3 B-parameter configuration (62.39\%). Further scaling to 0.8 B parameters yields almost no additional benefit (62.47\%), while incurring considerably higher latency, memory footprint, and training cost. We therefore adopt the 0.3 B DINOv3 backbone---used in GR-Pro---as our default, as it lies near the knee of the accuracy--efficiency trade-off.

\section{Future Research Opportunities}

We summarize our thoughts on future research opportunities inspired by our benchmarking results.

\paragraph{Multimodal and intent-conditioned retrieval.}
Our current instantiation of \textsc{LookBench} evaluates image-to-image retrieval, a core operation in deployed fashion systems, while practical search experiences increasingly mix screenshots, free-form text (``something like this but in navy''), structured attributes, and constraints such as budget or occasion.  Recent work on generalized multimodal fashion retrieval~\citep{zhu2024generalized} suggests that models should jointly encode images, language, and attribute signals, and reason about user intent rather than raw visual similarity alone. Extending retrieval formulations toward intent-conditioned, multi-source queries---for example, combining a reference look with textual edits and style preferences---is a natural next step for fashion-specific model design.

\paragraph{Richer fashion benchmarks and evaluation protocols.}
\textsc{LookBench} targets contamination-aware, realistic single- and multi-item image-to-image retrieval, but many aspects of fashion interaction remain underexplored on the evaluation side. Future benchmarks could incorporate temporal dynamics, e.g., trends across seasons~\citep{jin2023large}, user-level personalization~\citep{wang2023enhancing,chu2023leveraging}, and robustness tests under occlusion, cropping, or domain shifts between studio and social media imagery. Retrieval-augmented generation (RAG) settings, where a fashion assistant grounds its recommendations~\citep{xue2023weaverbird,xue2024demonstration}, explanations, or edits in a large, evolving catalog, also call for protocols that jointly assess retrieval quality and downstream generative behavior rather than retrieval in isolation.

\paragraph{Model design for aesthetic-aware fashion understanding.}
Beyond category and attribute correctness, fashion retrieval must capture subjective aesthetics such as silhouette balance, color harmony, and brand “mood.” Existing CLIP-based aesthetic predictors and LAION-style aesthetic scores offer scalable but coarse supervision that can be misaligned with human taste~\citep{schuhmann2022laion-aesthetics}. Recent work on aesthetic reasoning with multimodal LLMs shows that art-specific task decomposition and constrained chain-of-thought prompting can substantially improve alignment between model rankings and human preferences~\citep{Jiang_2025}. Adapting these ideas to fashion suggests retrieval backbones coupled with lightweight aesthetic reasoners that separate content (category, fit, function) from style (mood, formality, brand DNA) and act as plug-in rerankers or critics over candidate outfits, moving evaluation toward ``does this look good together?'' rather than only ``is this the same item?''.

\section{Conclusion}
We release \textsc{LookBench} together with all dataset splits, evaluation code, and baseline model outputs, and we design the benchmark to be periodically refreshed with new, time-stamped queries and corpora. This contamination-aware, fashion-specific setting is intended as a shared testbed for studying single and multi-item retrieval in both studio and street domains. We welcome community contributions for extending the tasks, adding new models and modalities, and maintaining \textsc{LookBench} as a living benchmark for fashion retrieval research.

\clearpage

\bibliographystyle{icml2020_url}
\bibliography{reference}

\clearpage
\appendix
\appendixpage

\crefalias{section}{appendix}
\crefalias{subsection}{appendix}

\section{Dataset Details}
\label{app:dataset}

\subsection{Query and Corpus Collections}
\label{app:query_collect}

\begin{figure}
    \centering
    \includegraphics[width=0.98\linewidth]{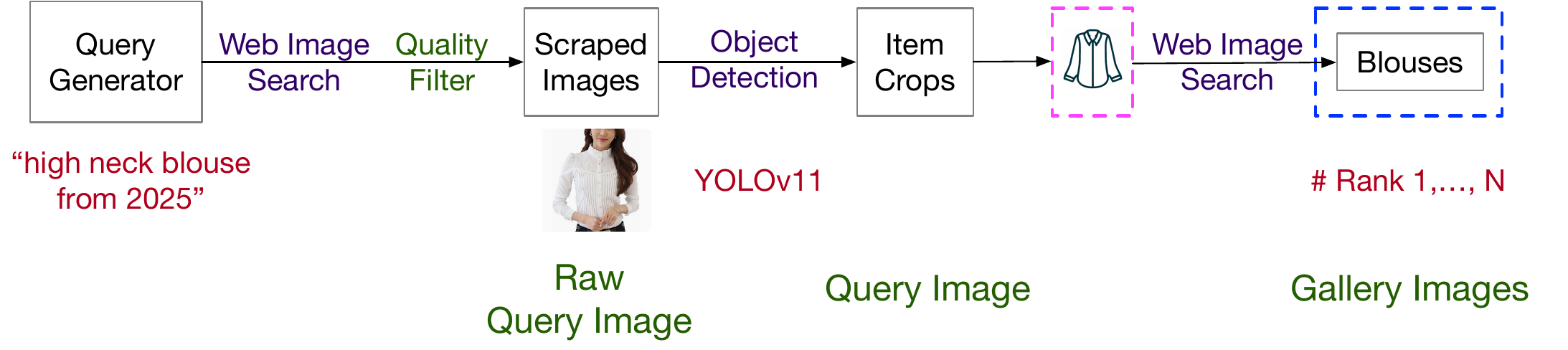}
    \caption{RealStudioFlat retrieval sets construction pipeline in \textsc{LookBench}.}
    \label{fig:data_collect_studio_flat}
\end{figure}

\paragraph{Query generation and ranked result selection for RealStudioFlat dataset.}
We construct the studio product portion of \textsc{LookBench}, RealStudioFlat, using the data collection pipeline illustrated in \cref{fig:data_collect_studio_flat}. Starting from our category–attribute taxonomy, we instantiate structured text templates (\Cref{lst:query_studio_flat}) to generate approximately 2{,}000 unique queries (e.g., ``high-neck blouse clothing product image, isolated on white background, from 2025''). Each text query is issued to an image search engine, and the returned results are time-stamped, de-duplicated, and filtered for quality to yield a pool of clean studio product photos—specifically flat-lay packshots.

We then apply YOLOv11 to localize garments within these images and generate category-labeled crops, treating each cropped region as a query image. For each crop, we filter the search results to retain only images that match the crop’s main attribute, which form the candidate pool for ranking. The number of candidates may vary depending on how many retrieved images satisfy this constraint. We then rank the candidates based on how many additional attributes they share with the query crop. The attribute annotation process is detailed in the following paragraphs. For evaluation, we retain up to the top-$k$ ranked results (ranks \#1 to $k$, if available) and store all crop–retrieved image pairs, along with associated metadata: the query string, timestamp, source URL, bounding box, category and attributes, and rank position. These pairs constitute the RealStudioFlat subset within \textsc{LookBench}.


\begin{lstlisting}[caption={Template for flat-lay studio product search queries.},label={lst:query_studio_flat}]
{category} {attribute} clothing product image, isolated on white background,
studio lighting, no model, from {year}.
\end{lstlisting}


\paragraph{Query generation and ranked result selection for RealStreetLook dataset.}
We construct retro lists from live street imagery using the pipeline in \cref{fig:data_collect}. Starting from our category–attribute taxonomy, we sample (category, attribute) combinations and instantiate the template in \Cref{lst:query_street_look} to form styling queries such as ``high-neck blouse street style fashion photo''. Each query is issued to a commercial image search engine (Google Images); the returned results are time-stamped, de-duplicated, and quality-filtered to obtain a pool of recent street photos. We then run YOLOv11 to localize fashion objects and produce category-labeled crops (e.g., top, skirt, shoe, bag).

For each detected crop, we follow the same candidate filtering and attribute-based ranking process previously described for the RealStudioFlat dataset. Specifically, we retain only images matching the crop’s main attribute and rank them by the number of additional shared attributes. Retrieved results are stored along with metadata—query string, timestamp, source URL, bounding box, category, attributes, and rank—forming the evaluation subset for RealStreetLook.

\begin{lstlisting}[caption={Template used to instantiate street-look search queries from the category–attribute taxonomy.},label={lst:query_street_look}]
{category} {attribute} street style fashion photo, model wearing, full body outfit, natural light, from {year}.
\end{lstlisting}

\paragraph{Query generation and ranked result selection for AIGen dataset.}
The AIGen subsets follow the same candidate filtering and attribute-based ranking procedure as described for the RealStudioFlat and RealStreetLook datasets. The key difference lies in the image source: instead of real-world web photos, the AIGen images are generated by a large multimodal generative model, Qwen2.5-VL-72B-Instruct. We use the same category–attribute taxonomy and prompt templates to condition image generation, ensuring stylistic and structural alignment with the real-data subsets. The resulting images are automatically time-stamped and stored along with their associated prompts. After generation, we apply YOLOv11 to detect fashion items, localize garments, and produce category-labeled crops. As in the real-image pipelines, each crop is used to retrieve and rank matching images within the AIGen corpus, based on attribute overlap. Retrieved results and metadata—query prompt, timestamp, bounding box, category, attributes, and rank—are stored as part of the AIGen-Studio and AIGen-StreetLook evaluation sets.

\paragraph{Query generation and ranked result selection for AIGen dataset.}
To generate synthetic images for the AIGen-Studio and AIGen-StreetLook subsets, we adopt a two-stage prompt-based pipeline using Qwen models.

In the first stage, we use Qwen2.5-VL-72B-Instruct to produce diverse natural language descriptions conditioned on a fashion concept defined by a (category, attribute) pair. For example, given a pair such as (shirt, v-neck), we randomly sample auxiliary parameters—such as lighting, background, pose, camera angle, location, and style—from curated option lists (see \cref{tab:aigen_placeholders}). These values are inserted into prompt templates (see \Cref{lst:prompt_aigen_studio} and \Cref{lst:prompt_aigen_streetlook}) to form visually rich instructions. This stage ensures semantic control while introducing stylistic and contextual variation across samples.

In the second stage, the generated textual description is used to condition the Qwen-Image-Edit~\citep{wu2025qwenimagetechnicalreport} to synthesize high-quality fashion imagery. To enhance visual fidelity, a fixed positive augmentation phrase (e.g., ``Ultra HD, 4K, cinematic composition'') is appended to the prompt. Images are generated at a resolution of 1664 $\times$ 928 pixels using consistent inference settings including guidance scale and fixed random seed. The output image is saved and time-stamped.

This two-stage process is repeated for ~2000 pairs of (category, attribute) combinations. By varying contextual parameters such as scene, lighting, and pose, we increase intra-class variability and make the retrieval task more challenging.

Once images are generated, we apply YOLOv11 to detect fashion items and extract category-labeled crops. These cropped regions serve as query images, and retrieval is performed using the same filtering and attribute-based ranking procedure used for the real image datasets.

\begin{table*}[ht]
  \centering
  \small
  \renewcommand{\arraystretch}{1.1}
  \setlength{\tabcolsep}{10pt}
  \begin{sc}
   \begin{tabularx}{\textwidth}{l X}
    \toprule
    Placeholder & Sample Values \\
    \midrule
    Location         & Tokyo crossing, Paris alley, cyberpunk street, city park \\
    Lighting         & Golden hour, studio lighting, harsh sun, soft shadows \\
    Weather          & Sunny, rainy, foggy, overcast, windy \\
    Style            & Street photography, fashion editorial, documentary \\
    Camera angle     & Eye-level, low angle, Dutch angle, close-up \\
    Time of day      & Early morning, golden hour, rush hour \\
    Background       & White, marble, pastel pink, wooden surface \\
    Layout           & On hanger, ghost mannequin, draped, upright on form \\
    Product angle    & Flat-lay, top-down, front view, isometric view \\
    Lighting (studio)& Soft studio, natural window light, backlit, rim lighting \\
    \bottomrule
  \end{tabularx}
  \end{sc}
\caption{Example placeholder values used for sampling visual diversity in AIGen prompt generation.}
\label{tab:aigen_placeholders}
\end{table*}

\Cref{lst:example_aigen_studio} and \Cref{lst:example_aigen_streetlook} are sample text prompts generated by Qwen2.5-VL-72B-Instruct and used to condition image generation via Qwen-Image. These prompts reflect diversity in scene context, lighting, angle, and styling while maintaining consistency with the (category, attribute) control inputs.

\begin{lstlisting}[caption={Template used to instantiate studio product prompts from the category–attribute taxonomy and sampled visual parameters.},label={lst:prompt_aigen_studio}]
Describe a clean, high-quality image of a {attribute} {category} displayed in full length. 
The item is {layout} against a {background} background. 
The shot is taken {angle}. Use {lighting}. 
Focus on the product details, material, and design.
\end{lstlisting}

\begin{lstlisting}[caption={Example of an instantiated studio prompt used for image generation.},label={lst:example_aigen_studio}]
Describe a clean, high-quality image of a v-neck shirt displayed in full length. 
The item is hanging on a hanger against a soft beige background. 
The shot is taken from a 45-degree angle. Use soft studio lighting. 
Focus on the product details, material, and design.
\end{lstlisting}

\begin{lstlisting}[caption={Template used to instantiate street-look prompts using diverse scene variables.},label={lst:prompt_aigen_streetlook}]
Create a detailed visual description for a {style} featuring a person wearing a '{attribute} {category}'. 
The setting is a {location} during {time} with {weather} weather and {lighting}. 
Capture the {angle}.
\end{lstlisting}

\begin{lstlisting}[caption={Example of an instantiated street-look prompt used for image generation.},label={lst:example_aigen_streetlook}]
Create a detailed visual description for a candid street photography scene featuring a person wearing a 'v-neck shirt'. 
The setting is a busy Tokyo crossing during golden hour with sunny weather and cinematic rim lighting. 
Capture the eye-level shot.
\end{lstlisting}

\paragraph{Corpus data collection}
Each retrieval subset in \textsc{LookBench} is paired with a retrieval corpus composed of two parts. The first part contains the ranked gallery results for every query, which are semantically aligned with the query crop in terms of garment category and key attributes, and constitute the positive target set.

To increase task difficulty and introduce realistic noise, we augment each corpus with additional distractor images drawn from Fashion200K. We compute CLIP ViT-L/14 embeddings for all \textsc{LookBench} query and gallery images, as well as for all Fashion200K images. For each task, we treat the union of its queries and gallery images as anchors and compute cosine similarity to every Fashion200K image. A Fashion200K image is retained as a distractor for that task if its similarity to at least one anchor falls within a task-specific band $[\alpha_t, \beta_t]$. This band removes near-duplicates ($>\beta_t$) and obviously unrelated items ($<\alpha_t$), yielding visually plausible but non-matching ``soft negatives'' that enrich the corpus. The similarity ranges and the resulting number of selected distractors are summarized in \cref{tab:corpus_noise}.

\begin{table*}[t]
  \centering
  \small
  \begin{sc}
  \begin{tabularx}{\textwidth}{l C C}
    \toprule
    Task & Similarity filter range & \# Noise images selected \\
    \midrule
    RealStudioFlat   & $[\,0.25, 0.45\,]$ & 58,275 \\
    AIGen-Studio     & $[\,0.25, 0.45\,]$ & 58,275 \\
    AIGen-StreetLook & $[\,0.25, 0.45\,]$ & 58,275 \\
    RealStreetLook   & $[\,0.25, 0.45\,]$ & 58,275 \\
    \bottomrule
  \end{tabularx}
  \end{sc}
  \caption{Similarity thresholds used to select Fashion200K distractor images for each \textsc{LookBench} task.}
  \label{tab:corpus_noise}
\end{table*}

\begin{table*}[t]
  \centering
  \small
  \renewcommand{\arraystretch}{1.10}
  \begin{sc}
  \begin{tabularx}{\textwidth}{l X}
    \toprule
    Category & Attributes \\
    \midrule
    shirt      & v-neck, crew neck, button-down, collared, long sleeve, short sleeve, slim fit, regular fit, loose fit, cotton, linen, denim, stripe, check, plain, oxford, pocket, formal, casual \\
    blouse     & v-neck, sweetheart, round neck, off the shoulder, long sleeve, puff sleeve, bishop sleeve, flared sleeve, silk, chiffon, satin, floral, lace, ruffled, pleated, feminine, elegant \\
    t-shirt    & crew neck, round neck, v-neck, short sleeve, oversized fit, regular fit, cotton, jersey, graphic, logo print, plain, street, casual \\
    sweatshirt & crew neck, hooded, zip-up, raglan sleeve, long sleeve, oversized, cotton, fleece, plain, printed, sporty, casual, street \\
    sweater    & crew neck, v-neck, turtle neck, high neck, knit, cable knit, ribbed, wool, cashmere, cotton, stripe, plain, cozy, winter, casual \\
    cardigan   & v-neck, button, open front, long sleeve, knit, wool, cropped, longline, plain, stripe, soft, cozy, casual \\
    jacket     & zip-up, button, biker, bomber, trucker, cropped, denim, leather, suede, wool, plain, camouflage, sporty, street, casual \\
    vest       & v-neck, crew neck, sleeveless, button, zip-up, denim, knit, leather, puffer, plain, sporty, casual \\
    pants      & high waist, mid waist, low waist, straight leg, slim fit, wide leg, flared, cargo, pleated, cropped, denim, cotton, wool, plain, check, formal, casual, street \\
    shorts     & high waist, mid waist, denim, cotton, cargo, pleated, paperbag, belted, plain, check, casual, beach \\
    skirt      & high waist, a-line, pencil, pleated, flared, mini length, midi length, maxi length, denim, cotton, silk, floral, plain, elegant, feminine, casual \\
    coat       & long sleeve, single breasted, double breasted, trench, pea coat, overcoat, wool, cashmere, leather, plain, check, formal, elegant, winter \\
    dress      & v-neck, round neck, off the shoulder, halter, short sleeve, long sleeve, sleeveless, fit and flare, a-line, bodycon, wrap, mini length, midi length, maxi length, floral, lace, silk, chiffon, elegant, casual \\
    jumpsuit   & v-neck, round neck, sleeveless, short sleeve, wide leg, straight leg, belted, denim, cotton, plain, floral, casual, modern \\
    cape       & a-line, oversized, wool, cashmere, plain, check, elegant, formal \\
    glasses    & metal, plastic, round, square, cat eye, aviator, retro, modern \\
    hat        & wool, cotton, felt, plain, floral, bow, brimmed, beanie, casual, elegant \\
    headband   & plain, floral, ribbon, bow, pearl, sequin, decorative, feminine \\
    tie        & plain, stripe, geometric, silk, knit, formal, business \\
    glove      & leather, suede, wool, plain, quilted, winter, elegant \\
    watch      & metal, leather, round, square, minimalist, sport, luxury \\
    belt       & leather, suede, metal buckle, chain, plain, casual, formal \\
    sock       & cotton, wool, ankle, crew, knee-high, plain, stripe, dot, casual \\
    shoe       & leather, suede, rubber, lace-up, slip-on, buckle, flat, heel, boot, sneaker, formal, casual, sporty \\
    bag        & leather, suede, crocodile, snakeskin, tote, crossbody, clutch, backpack, bead, sequin, plain, elegant, casual \\
    scarf      & silk, wool, cashmere, fringe, floral, stripe, plain, elegant, winter \\
    umbrella   & plain, stripe, geometric, folding, transparent, compact \\
    \bottomrule
  \end{tabularx}
  \end{sc}
  \caption{Category–attribute taxonomy used for fine-grained fashion descriptions in \textsc{LookBench}.}
  \label{tab:attr_taxonomy}
\end{table*}

\begin{figure}
    \centering
    \includegraphics[width=0.95\linewidth]{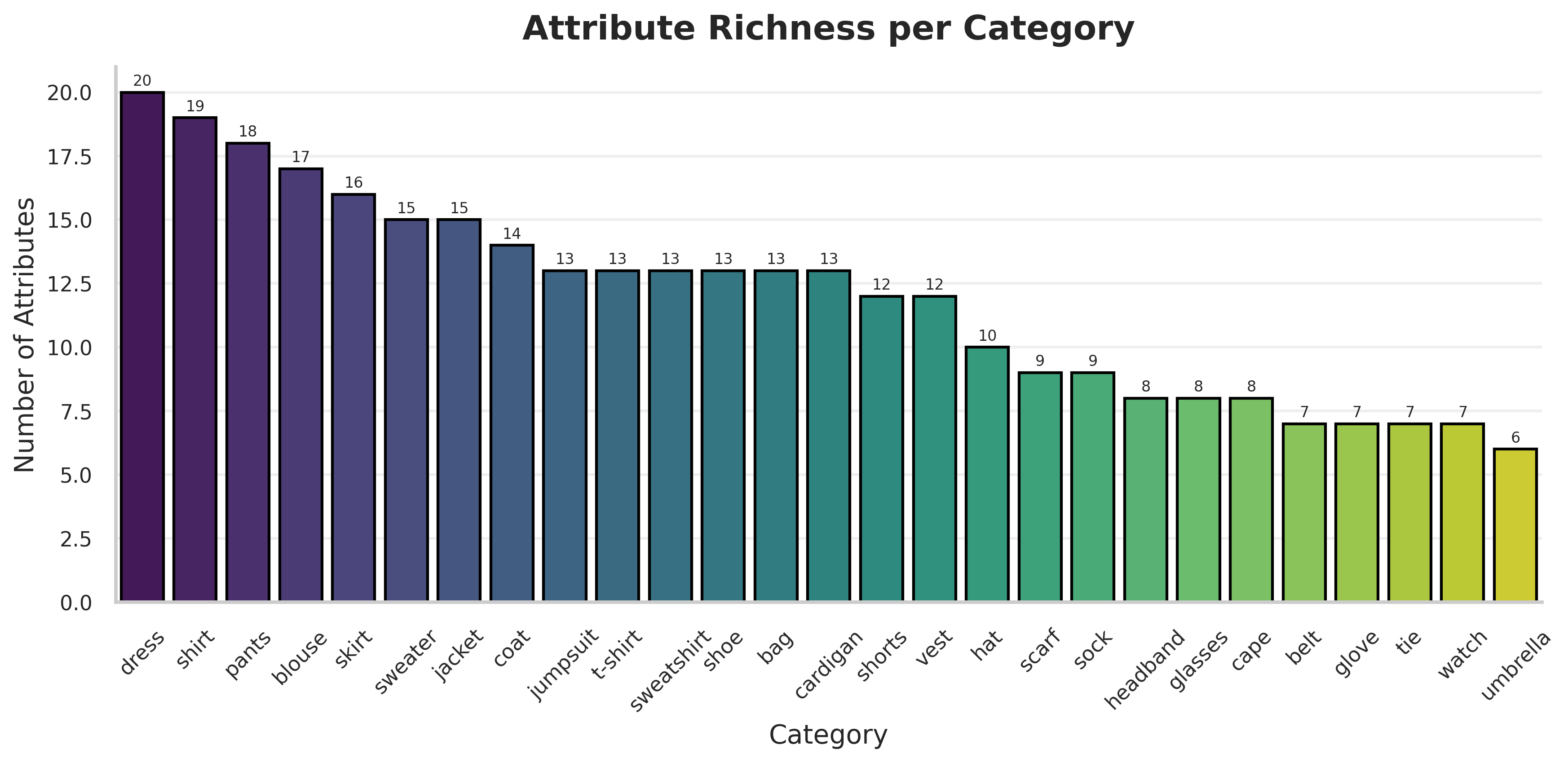}
    \caption{Attribute taxonomy used in \textsc{LookBench}. 
Each bar shows the number of visually grounded attributes defined for a given fashion category, illustrating that core apparel types (e.g., dress, shirt, pants) are described with the richest attribute sets.
}
    \label{fig:att_distribution}
\end{figure}

\subsection{Attribute taxonomy and pre-annotation.}
\label{app:annotation}

\paragraph{Attribute taxonomy.}
We define a fine-grained category–attribute taxonomy to support visually grounded fashion retrieval. Each fashion category (e.g., \textit{shirt}, \textit{dress}, \textit{bag}) is associated with a curated list of 10–25 attributes that are both visually discernible and semantically meaningful (see \cref{tab:attr_taxonomy}). These attributes capture garment structure (e.g., \textit{v-neck}, \textit{button-down}), material and texture (e.g., \textit{wool}, \textit{lace}), fit and cut (e.g., \textit{slim fit}, \textit{oversized}), and style intent (e.g., \textit{elegant}, \textit{street}). 

Unlike prior fashion benchmarks that use fixed global tags or noisy keyword extractions, our taxonomy is:
\begin{itemize}[leftmargin=*]
  \item Category-specific: each attribute list is tailored to a single category, improving label relevance and reducing ambiguity.
  \item Visual-first: attributes are selected based on visual detectability in images, rather than metadata or text descriptions.
  \item Discriminative: the attributes are designed to support fine-grained visual comparisons, enabling retrieval of subtle variations (e.g., distinguishing \textit{pleated} vs. \textit{ruffled} skirts).
\end{itemize}

\paragraph{Attribute pre-annotation.}
To obtain fine-grained category–attribute labels at scale, we pre-annotate item crops using Qwen2.5-VL-72B. For each detected garment region, we provide the image crop, its predicted category, and the corresponding attribute vocabulary from our taxonomy, and prompt the model to identify only attributes that are visually supported. The model returns a JSON object containing one \texttt{main\_attribute}---the most visually salient feature---and a list of \texttt{other\_attributes} that are also clearly visible. A sample prompt is shown in Listing~\Cref{lst:attr_annot_prompt}. This procedure is run offline across all item crops in the benchmark.

\begin{lstlisting}[caption={Prompt template for Qwen2.5-VL-72B attribute pre-annotation.},label={lst:attr_annot_prompt}]
System: You are an expert fashion stylist and visual tagger.
You see an image crop of a single fashion item.

User:
You are given:
1) The target category: {category}
2) A list of allowed attributes for this category:
   {attribute_vocabulary}

Your task:
- Focus only on the target item in the crop (ignore background and other items).
- Identify one main attribute that is most visually prominent - this goes under "main_attribute".
- Also list all other attributes from the vocabulary that are clearly visible -these go under "other_attributes".
- Do NOT invent new attributes and do NOT use words outside the list.
- If no attributes are visible or confident, leave both fields empty.

Return your answer in strict JSON format:

{
  "category": "{category}",
  "main_attribute": "attr_x",
  "other_attributes": [
    "attr_1",
    "attr_2",
    ....
  ]
}
\end{lstlisting}

\paragraph{Validation.}
To evaluate the quality of our attribute annotations, we adopt a LLM-based verification protocol using GPT-5.1 in an \emph{LLM-as-a-judge} framework. This setup enables consistent and reproducible assessments of visual attribute correctness across diverse item types and styles, without requiring manual labeling.

For each sampled image crop, GPT-5.1 is given:
\begin{itemize}[leftmargin=*]
  \item The input image crop showing the target item.
  \item Its assigned fashion category.
  \item The category-specific attribute vocabulary from our taxonomy.
  \item The predicted attributes from Qwen2.5-VL-72B, including both the \texttt{main\_attribute} and the list of \texttt{other\_attributes}.
\end{itemize}

The model is instructed to verify each predicted attribute independently, based solely on visual evidence in the image. A sample prompt is shown in \Cref{lst:attr_judge_prompt}.

We randomly sample 200 item crops across the dataset and apply this judging procedure offline. We aggregate all predicted attributes (main + other) and compute correctness using the formula:
\[
\text{Correctness} =
\frac{\#\text{attributes judged correct}}{\#\text{attributes predicted in total}}.
\]
Across the 200-sample evaluation set, we observe an estimated correctness rate of approximately 93\%. Given this high precision, we treat the pre-annotations as a reliable source of weak supervision for downstream retrieval evaluation, attribute analysis, and model benchmarking.

\begin{lstlisting}[caption={Prompt template for GPT-5.1 LLM-as-a-judge evaluation of attribute annotations.},label={lst:attr_judge_prompt}]
System: You are a precise fashion evaluation assistant.
You will judge whether a set of attributes correctly describes
a clothing item in an image crop.

User:
You are given:
1) The target category: {category}
2) The list of allowed attributes for this category:
   {attribute_vocabulary}
3) The model-predicted attributes for this image:
   {
     "main_attribute": "attr_x",
     "other_attributes": ["attr_1", "attr_2", ...]
   }
4) An image crop showing the target item.

Your task:
- For EACH predicted attribute (main + others), decide whether it is:
    - "correct"   : clearly supported by the image
    - "incorrect" : clearly NOT supported by the image
- Base your decision ONLY on visible details of the target item.
- Do NOT add new attributes or explanations.

Return your answer in strict JSON format:

{
  "per_attribute": [
    {"name": "attr_x", "verdict": "correct"},
    {"name": "attr_1", "verdict": "correct"},
    {"name": "attr_2", "verdict": "incorrect"}
  ]
}
\end{lstlisting}

\subsection{Fashion200K Evaluation Set Construction}
We construct the evaluation set from the publicly available Fashion200K dataset released by Marqo\footnote{\small \url{https://huggingface.co/datasets/Marqo/fashion200k}}. Starting from the full corpus, we encode all images using a CLIP visual encoder and perform large-scale clustering in the embedding space to group visually similar items. From this process, we form 50,000 clusters, each representing a distinct visual concept or product style. To ensure high diversity and avoid near-duplicate queries, we randomly select a single representative image from each cluster, resulting in a query set of 50,000 images. This clustering-based sampling strategy promotes broad coverage of visual attributes such as category, color, texture, and silhouette, making the benchmark a robust evaluation of large-scale image retrieval performance.

\section{Experiment Details}
\label[appendix]{app:exp}

\subsection{Implementation Details}\label{app:imp}
For the analysis and data processing framework, we use the code from  the  public Github  repository at {\small \url{https://github.com/wuji3/Doraemon}}~\citep{du2025visual} with GNU General Public License.

For the evaluation framework, we use the code from  the  public Github  repository at {\small \url{https://github.com/beir-cellar/beir}}~\citep{thakur2021beir} with Apache License and   {\small \url{https://github.com/eosphoros-ai/DB-GPT-Hub}}~\citep{xue2023dbgpt,zhou2024dbgpthub} with MIT License.

\subsection{Training and Testing Details}\label{app:training_details}

To benchmark our method against a broad spectrum of existing models, we evaluate models with input resolutions ranging from 224 to 448 and feature dimensionalities spanning 512 to 4096. Although these models differ in architecture and representation size, all evaluations are conducted on a single GPU with the batch size fixed at 128 to ensure consistent and efficient comparison. For a comprehensive assessment of retrieval performance, we report Recall@1, MRR@5, and nDCG.


\subsection{Evaluation Metrics}
\label[appendix]{app:metrics}

\paragraph{Fine and coarse Recall@k.}
Recall is a commonly used metric in retrieval, measuring the proportion of relevant items successfully retrieved within the top-$k$ results. In \textsc{LookBench}, relevance is defined with respect to our category--attribute taxonomy.

For a query with garment category $c_q$ and attribute set $A_q$, and a retrieved item at position $i$ with category $c_i$ and attribute set $A_i$, we define two binary relevance indicators:
\begin{align}
rel_{\text{coarse}}(i) &= 
\begin{cases}
1, & \text{if } c_i = c_q,\\
0, & \text{otherwise},
\end{cases} \\
rel_{\text{fine}}(i) &=
\begin{cases}
1, & \text{if } c_i = c_q \text{ and } A_q \subseteq A_i,\\
0, & \text{otherwise},
\end{cases}
\end{align}
where $rel_{\text{coarse}}$ tests only for category consistency, and $rel_{\text{fine}}$ requires the retrieved item to match the category and contain all annotated attributes of the query.

Given $R_{\text{coarse}}$ and $R_{\text{fine}}$ relevant items in the corpus under the corresponding notion of relevance, we compute:
\begin{align}
\mathrm{CoarseRecall@}k &= \frac{1}{R_{\text{coarse}}} \sum_{i=1}^{k} rel_{\text{coarse}}(i), \\
\mathrm{FineRecall@}k   &= \frac{1}{R_{\text{fine}}}   \sum_{i=1}^{k} rel_{\text{fine}}(i).
\end{align}
Both metrics lie in $[0,1]$ and are non-decreasing with $k$. Fine Recall@1, our main metric in the paper, is thus a strict indicator of whether the top-ranked item is an attribute-perfect match to the query.

\medskip
\paragraph{nDCG@k.}
Normalized Discounted Cumulative Gain (nDCG) evaluates the quality of the ordering of retrieved items using graded (non-binary) relevance. For the same query $(c_q, A_q)$ and item $(c_i, A_i)$, we define a real-valued relevance score that rewards both category correctness and attribute overlap:
\begin{equation}
rel(i) = \mathbf{1}[c_i = c_q] \cdot \frac{|A_q \cap A_i|}{|A_q|},
\end{equation}
so that $rel(i) \in [0,1]$, with higher values indicating more attribute-consistent matches among items of the correct category.

The discounted cumulative gain at rank $k$ (DCG@k) is then:
\begin{equation}
\mathrm{DCG@}k = \sum_{i=1}^{k} \frac{rel(i)}{\log_2(i+1)}.
\label{eq:dcg}
\end{equation}
The ideal DCG (IDCG@k) is obtained by sorting the retrieved items in descending order of $rel(i)$ and taking:
\begin{equation}
\mathrm{IDCG@}k = \sum_{i=1}^{k} \frac{rel^{\star}(i)}{\log_2(i+1)},
\label{eq:idcg}
\end{equation}
where $rel^{\star}(i)$ is the $i$-th largest relevance score achievable for the query. Finally, nDCG@k is defined as:
\begin{equation}
\mathrm{nDCG@}k = \frac{\mathrm{DCG@}k}{\mathrm{IDCG@}k},
\label{eq:ndcg}
\end{equation}
which ensures $\mathrm{nDCG@}k \in [0,1]$. In the main paper we focus on nDCG@5.

\medskip
\paragraph{MRR.}
Mean Reciprocal Rank (MRR) measures the position of the first relevant retrieved item for each query. For a set of $Q$ queries, it is computed as:
\begin{equation}
\mathrm{MRR} = \frac{1}{Q} \sum_{q=1}^{Q} \frac{1}{\mathrm{rank}(q)},
\label{eq:mrr}
\end{equation}
where $\mathrm{rank}(q)$ is the position of the first (fine-)relevant item for query $q$; if no relevant item appears in the list, we set $\frac{1}{\mathrm{rank}(q)} = 0$. Although our evaluation code reports MRR for all models, we omit detailed MRR tables from the paper due to space constraints, as the trends closely mirror those of Fine Recall@k.

\subsection{More Experiment Results}
\label{app:exp}

\paragraph{Results measured in other metrics.}  The results measured in coarse Recall@1 are in \cref{tab:coarse_clean} while the results in nDCG@5 in \cref{tab:ndcg_clean}.

\begin{table*}[tb]
  \centering
  \small
  \renewcommand{\arraystretch}{1.12}
  \setlength{\tabcolsep}{3.5pt}
  \begin{sc}
  \begin{tabularx}{\textwidth}{l  *{1}{l}*{4}{S}*{1}{S}}
    \toprule
    Model
      & Resolution
      & \multicolumn{2}{c}{AIgen}
      & \multicolumn{2}{c}{Real}
      & Overall \\
      \cmidrule(lr){3-4} \cmidrule(lr){5-6}
      & / Emb. Size & StreetLook & Studio & StreetLook & Studio &  \\
    \midrule
    \rowcolor{aigreen}%
    GR-Pro          & 336 / 1024 & \textbf{63.67} & \textbf{54.88} & \textbf{44.75} & \textbf{51.55} & \textbf{49.80} \\
    \rowcolor{aigreen!70}%
    GR-Lite         & 336 / 1024 & 62.47 & 52.08 & 43.84 & 51.70 & 49.18 \\
    Marqo-fashionCLIP    & 224 / 512  & 63.22 & 54.93 & 41.87 & 51.68 & 48.63 \\
    Marqo-fashionSigLIP  & 224 / 768  & 66.27 & 58.53 & 42.43 & 51.86 & 49.44 \\
    SigLIP2-B/16         & 384 / 768  & 57.83 & 54.97 & 39.35 & 49.12 & 46.10 \\
    SigLIP2-L/16         & 384 / 1024 & 51.89 & 48.57 & 35.91 & 44.78 & 41.86 \\
    PP-ShiTuV2           & 224 / 512  & 30.06 & 33.69 & 32.77 & 43.22 & 37.17 \\
    DINOv3-ViT-L         & 224 / 1024 & 20.24 & 27.66 & 26.27 & 39.85 & 31.83 \\
    DINOv3-ViT-7B        & 224 / 4096 & 19.42 & 30.30 & 26.77 & 39.38 & 31.99 \\
    DINOv2-ViT-L         & 224 / 1024 & 24.29 & 25.05 & 22.99 & 37.66 & 29.57 \\
    DINOv2-ViT-G         & 224 / 1536 & 23.13 & 23.27 & 22.96 & 37.71 & 29.36 \\
    CLIP-L/14            & 336 / 768  & 25.28 & 25.95 & 21.09 & 40.35 & 30.08 \\
    DINOv3-ConvNext      & 224 / 1536 & 10.45 & 19.13 & 19.12 & 33.38 & 24.68 \\
    CLIP-B/16            & 224 / 512  & 17.86 & 13.75 & 16.80 & 34.75 & 24.36 \\
    InternViT-6B         & 448 / 1024 & 14.17 & 20.13 & 16.40 & 31.33 & 22.99 \\
    \bottomrule
  \end{tabularx}
  \end{sc}
  \caption{
    Comparison of nDCG@5 across the four \textsc{LookBench} subsets.
    ``Overall'' is a query-count-weighted average. Best values per column are shown in \textbf{bold}.
  }
  \label{tab:ndcg_clean}
\end{table*}

\begin{table*}[tb]
  \centering
  \small
  \renewcommand{\arraystretch}{1.12}
  \setlength{\tabcolsep}{3.5pt}
  \begin{sc}
  \begin{tabularx}{\textwidth}{l  *{1}{l}*{4}{S}*{1}{S}}
    \toprule
    Model
      & Resolution
      & \multicolumn{2}{c}{AIgen}
      & \multicolumn{2}{c}{Real}
      & Overall \\
      \cmidrule(lr){3-4} \cmidrule(lr){5-6}
      & / Emb. Size & StreetLook & Studio & StreetLook & Studio &  \\
    \midrule
    \rowcolor{aigreen}%
    GR-Pro          & 336 / 1024 & \textbf{92.50} & \textbf{92.75} & \textbf{79.82} & \textbf{94.16} & \textbf{87.93} \\
    \rowcolor{aigreen!70}%
    GR-Lite         & 336 / 1024 & 88.75 & 90.16 & 76.76 & 92.68 & 85.54 \\
    Marqo-fashionCLIP    & 224 / 512  & 84.38 & 87.05 & 75.33 & 88.72 & 82.68 \\
    Marqo-fashionSigLIP  & 224 / 768  & 90.00 & 93.78 & 73.39 & 88.63 & 82.77 \\
    SigLIP2-B/16         & 384 / 768  & 86.25 & 90.67 & 72.17 & 88.33 & 81.62 \\
    SigLIP2-L/16         & 384 / 1024 & 80.62 & 90.67 & 68.20 & 84.97 & 78.12 \\
    PP-ShiTuV2           & 224 / 512  & 51.88 & 62.18 & 60.04 & 78.83 & 67.76 \\
    DINOv3-ViT-L         & 224 / 1024 & 40.00 & 64.25 & 56.68 & 78.73 & 65.67 \\
    DINOv3-ViT-7B        & 224 / 4096 & 33.12 & 69.95 & 56.07 & 78.04 & 65.12 \\
    DINOv2-ViT-L         & 224 / 1024 & 51.25 & 67.36 & 55.86 & 79.23 & 66.57 \\
    DINOv2-ViT-G         & 224 / 1536 & 51.25 & 60.62 & 54.84 & 79.13 & 65.55 \\
    CLIP-L/14            & 336 / 768  & 46.88 & 56.48 & 45.26 & 76.85 & 59.91 \\
    DINOv3-ConvNext      & 224 / 1536 & 30.63 & 49.22 & 43.53 & 74.38 & 56.42 \\
    CLIP-B/16            & 224 / 512  & 35.62 & 32.12 & 33.54 & 67.26 & 48.11 \\
    InternViT-6B         & 448 / 1024 & 25.62 & 45.60 & 34.45 & 60.83 & 46.14 \\
    \bottomrule
  \end{tabularx}
  \end{sc}
  \caption{
    Comparison of \textsc{Coarse-Recall@1} across four \textsc{LookBench} subsets.
    ``Overall'' is a query-count-weighted average. Best values per column are shown in \textbf{bold}.
  }
  \label{tab:coarse_clean}
\end{table*}

\section{Licensing of DINOv3 and GR-Lite}
\label{app:licensing}

\paragraph{Upstream DINOv3 license.}
Our GR-Lite model is obtained by fine-tuning the publicly released DINOv3
vision encoder.\footnote{DINOv3 repository and license: \small \url{https://github.com/facebookresearch/dinov3}.}
DINOv3 is distributed by Meta under the ``DINOv3 License''. In summary, this license grants us a
non-exclusive, worldwide, royalty-free license to use, reproduce, distribute,
copy, create derivative works of, and make modifications to the DINOv3 models,
software, and accompanying documentation (collectively, the ``DINO Materials''),
subject to several conditions.

\paragraph{GR-Lite as a derivative work of DINOv3.}
GR-Lite is obtained by initializing from the DINOv3 vision backbone and
fine-tuning it on an open-source fashion corpus using our attribute-supervised
objectives (see \cref{sec:training}). As such, the GR-Lite model weights
constitute a derivative work of the DINO Materials in the sense of the DINOv3
License. In accordance with this license:
\begin{itemize}[leftmargin=*]
    \item We distribute GR-Lite's pretrained weights under the same DINOv3
    License terms, and include a copy of the license with the released model.
    \item We clearly acknowledge the use of DINOv3 in this paper and in the
    accompanying code and model release.
    \item To the extent our code, documentation, or training scripts are our
    own original work and do not contain DINO Materials, we may license those
    components under separate open-source terms (e.g., MIT/Apache), but this
    does not alter the fact that the GR-Lite weights themselves remain subject
    to the DINOv3 License.
\end{itemize}

\end{document}